\documentclass{IEEEtran} 
\usepackage{helvet}
\usepackage{courier}
\usepackage[latin9]{inputenc}
\usepackage{amsmath}
\usepackage{amssymb}
\usepackage[final]{graphicx}
\usepackage{subcaption}
\usepackage{esint}
\makeatletter
\usepackage{atbegshi}% http://ctan.org/pkg/atbegshi
\AtBeginDocument{\AtBeginShipoutNext{\AtBeginShipoutDiscard}}

\DeclareMathOperator*{\argmax}{arg\,max}
\pdfoutput=1

\usepackage{color}
\usepackage{algorithm}
\usepackage{caption}
\usepackage{algcompatible}
\usepackage{amsfonts}
\usepackage{amsthm}
\usepackage{cite}
\usepackage{multirow}
\usepackage{color}

\newtheorem{theorem}{Theorem}

\newtheorem{assumption}{Assumption}
\newtheorem{definition}{Definition}

\newtheorem{remark}{Remark}
\newtheorem{lemma}{Lemma}
\newtheorem{proposition}{Proposition}

\newcommand{\comment}[1]{}

\ifodd 1

\else

\fi

\ifodd 1

\else

\fi

\ifodd 1
\newcommand{\dtl}[1]{{\color{red}#1}}
\else
\newcommand{\dtl}[1]{#}
\fi

\ifodd 1
\newcommand{\quest}[1]{{\color{red}#1}}
\else
\newcommand{\quest}[1]{#}
\fi

\addtocounter{page}{-1}

\ifodd 1
\newcommand{\com}[1]{\textbf{\color{red}(COMMENT: #1)}} 

\else
\newcommand{\com}[1]{}

\fi

\begin{document}
% If your paper is accepted and the title of your paper is very long,% the style will print as headings an error message. Use the following% command to supply a shorter title of your paper so that it can be% used as headings.%\runningtitle{I use this title instead because the last one was very long}

% If your paper is accepted and the number of authors is large, the% style will print as headings an error message. Use the following% command to supply a shorter version of the authors names so that% they can be used as headings (for example, use only the surnames)%\runningauthor{Surname 1, Surname 2, Surname 3, ...., Surname n}

%\twocolumn[

\title{Global Bandits}
\thanks{
O. Atan and M. van der Schaar are with the Department of Electrical Engineering, University of California, Los Angeles, CA 90095 USA (e-mail:
oatan@ucla.edu; mihaela@ee.ucla.edu) C. Tekin is with the Department of Electrical and Electronics Engineering, Bilkent University, Ankara 06800, Turkey (e-mail: cemtekin@ee.bilkent.edu.tr). }
\author{
    \IEEEauthorblockN{Onur Atan, Cem Tekin, \textit{Member}, IEEE, Mihaela van der Schaar, \textit{Fellow}, IEEE \\ }
}

\maketitle


%\address{ Unknown Institution 1 and Unknown Institution 2 and Unknown Institution 3 } %] 

\begin{abstract}
Multi-armed bandits (MAB) model sequential decision making problems, in which a learner sequentially chooses arms with unknown reward distributions in order to maximize its cumulative reward. Most of the prior work on MAB assumes that the reward distributions of each arm are independent. But in a wide variety of decision problems -- from drug dosage to dynamic pricing -- the expected rewards of different arms are correlated, so that selecting one arm provides information about the expected rewards of other arms as well.  We propose and analyze a class of models of such decision problems, which we call {\em global bandits}. In the case in which rewards of all arms are deterministic functions of a single unknown parameter, we construct a greedy policy that achieves {\em bounded regret}, with a bound that depends on the single true parameter of the problem.  Hence, this policy selects suboptimal arms only finitely many times with probability one. For this case we also obtain a bound on regret that is {\em independent of the true parameter}; this bound is sub-linear, with an exponent that depends on the informativeness of the arms. We also propose a variant of the greedy policy that achieves $\tilde{\mathcal{O}}(\sqrt{T})$ worst-case and $\mathcal{O}(1)$ parameter dependent regret. Finally, we perform experiments on dynamic pricing and show that the proposed algorithms achieve significant gains with respect to the well-known benchmarks. 
\end{abstract}

\section{Introduction}

Multi-armed bandits (MAB) provide powerful models and algorithms for sequential decision-making problems in which the expected reward of each arm (action) is unknown. The goal in MAB problems is to design online learning algorithms that maximize the total reward, which turns out to be equivalent to minimizing the regret, where the regret is defined as the difference between the total expected reward obtained by an oracle that always selects the best arm based on complete knowledge of arm reward distributions, and that of the learner, who does not know the expected arm rewards beforehand.
Classical $K$-armed MAB \cite{lairobbinsl} does not impose any dependence between the expected arm rewards.  But in a wide variety of decision problems -- from drug dosage to dynamic pricing -- the expected rewards of different arms are correlated, so that selecting one arm provides information about the expected rewards of other arms as well. In this paper we propose and analyze such a MAB model, which we call {\em Global Bandits} (GB). 

In GB, the expected reward of each arm is a function of a single global parameter. It is assumed that the learner knows these functions but does not know the true value of the parameter. For this problem, we propose a greedy policy, which constructs an estimate of the global parameter by taking a weighted average of parameter estimates computed separately from the reward observations of each arm.
Then, we show that this policy achieves {\em bounded regret}, where the bound depends on the value of the parameter.
This implies that the greedy policy learns the optimal arm, i.e., the arm with the highest expected reward, in finite time. 
We also obtain a worst-case (parameter independent) bound on the regret of the greedy policy. We show that this bound is sub-linear in time, and its time exponent depends on the {\em informativeness of the arms}, which is a measure of the strength of correlation between expected arm rewards. 

GBs encompass the model studied in \cite{Tsiklis_structured}, in which it is assumed that the expected reward of each arm is a {\em linear function} of a single global parameter. This is a special case of the more general model we consider in this paper, in which the expected reward of each arm is a H\"{o}lder continuous, possibly non-linear function of a single global parameter. 
On the technical side, non-linear expected reward functions significantly complicates the learning problem. 
When the expected reward functions are linear, then the information one can infer about the expected reward of arm $X$ by an additional single sample of the reward from arm $Y$ is independent of the history of previous samples from arm $Y$.\footnote{The additional information about the expected reward of arm $X$ that can be inferred from obtaining sample reward $r$ from arm $Y$ is the same as the additional information about the expected reward of arm $X$ that could be inferred from obtaining the sample reward $L(r)$ from arm $X$ itself, where $L$ is a linear function that depends only on the reward functions themselves.} However, if reward functions are non-linear, then the additional information that can be inferred about the expected reward of arm $X$ by a single sample of the reward from arm $Y$ is biased. Therefore, the previous samples from arm $X$ and arm $Y$ needs to be incorporated to ensure that this bias asymptotically converges to $0$.

Many applications can be formalized as GBs. Examples include: (i) clinical trials involving similar drugs (e.g., drugs with a similar chemical composition) or treatments which may have similar effects on the patients; (ii) dynamic pricing with the objective of maximizing revenue over a finite time horizon.

\textbf{Example 1:} Let $y_t$ be the dosage level of the drug for patient $t$ and $x_t$ be the response of patient $t$. The relationship between the drug dosage and patient response is modeled in \cite{lai1978adaptive} as $x_t = M(y_t; \theta_{*}) + \epsilon_t$, where $M(\cdot)$ is the response function, $\theta_{*}$ is the slope if the function is linear or the elasticity if the function is exponential or logistic, and $\epsilon_t$ is i.i.d. zero mean noise. For this model, $\theta_{*}$ becomes the global parameter and the set of drug dosage levels becomes the set of arms. 

\textbf{Example 2:} In dynamic pricing, an agent sequentially selects a price from a finite set of prices ${\cal P}$ with the objective of maximizing its revenue over a finite time horizon \cite{dynamic}. At instance $t$, the agent first selects a price $p_t \in {\cal P}$, and then observes the amount of sales at time $t$, which is denoted by $S(p_t; \theta_{*})$. We have
$
S(p_t; \theta_{*}) = F(p_t; \theta_{*}) + \epsilon_t,
$
where $F(.)$ is the modulating function, $\theta_{*}$ is the market size and $\epsilon_{t}$ is the noise term with zero mean. The modulating function is equal to the purchase probability of an item of price $p_t$ given the market size $\theta_{*}$. Examples of commonly used modulating functions can be found in \cite{huang2013demand}. The revenue is then given by $R(p_t;\theta_{*}) = p_t F(p_t; \theta_{*}) + p_t \epsilon_t$.
In this example, the market size is the unknown global parameter which needs to be learned online by setting prices and observing the related revenues. In Section \ref{sec:numerical}, we illustrate the use of methods proposed in this paper on this dynamic pricing example. 

In addition to the above examples, GBs can also be applied in any setting in which the parameters of a system that depends on the rewards in a non-linear way need to be estimated in order to learn the optimal arms. At this point, it is important to note that our work differs from the existing works on non-linear parameter estimation \cite{li2008asymptotic, pakrooh2013analysis, iltis1999density} because its focus is to maximize the total reward by using the estimates of the parameter to decide which arms to select.

The remainder of the paper is organized as follows. Contribution and the key results are summarized in Section \ref{sec:contributions}. Related work is discussed in Section \ref{sec:relatedwork}. Problem formulation is given in Section \ref{sec:probdesc}. A greedy policy is proposed in Section \ref{sec:WAGP}, and its regret is analyzed in Section \ref{sec:WAGPregret}. An improved algorithm that combines the greedy policy with an upper confidence bound policy is proposed in Section \ref{sec:BUW}. 
Learning under time varying global parameter is considered in Section \ref{sec:ext}. 
Numerical results are given in Section \ref{sec:numerical}, followed by the concluding remarks given in Section \ref{sec:conclusion}. All proofs are given in the Appendix

\section{Contribution and Key Results}\label{sec:contributions}

This paper is an extended version of~\cite{atan2015}, adding the following contributions. First, it provides two new theoretical results on WAGP: mean-squared convergence of the estimated global parameter and a lower bound on the regret. Second, it provides two new algorithms: (i) BUW which switches between the UCB1 and WAGP in order to achieve optimal parameter dependent and worst-case regrets, (ii) non-stationary WAGP, which tracks the time varying global parameter to take optimal actions. Third, it provides an illustration of the use of the proposed algorithms on the dynamic pricing example. In addition, this paper has extended introduction and related work sections, and includes proofs of all theorems. Our main contributions can be summarized as follows: 
\begin{itemize}
\item We propose a non-linear parametric model for MABs, which we refer to as GBs, and a greedy policy, referred to as {\em Weighted Arm Greedy Policy} (WAGP), which achieves bounded regret.
\item We define the concept of {\em informativeness}, which measures how well one can estimate the expected reward of an arm by using rewards observed from the other arms, and then, prove a sublinear in time worst-case regret bound for WAGP that depends on the informativeness.
\item We also propose another learning algorithm called the {\em Best of UCB and WAGP (BUW)}, which fuses the decisions of the UCB1~\cite{A2002} and WAGP in order to achieve $\tilde{\mathcal{O}}(\sqrt{T})$\footnote{$\mathcal{O}(\cdot)$ is the Big O notation, $\tilde{\mathcal{O}}(\cdot)$ is the same as $\mathcal{O}(\cdot)$ except it hides terms that have polylogarithmic growth.} worst-case and $\mathcal{O}(1)$ parameter dependent regrets.
\item We study a non-stationary version of GB, where the global parameter slowly changes over time.
For this case, we prove a bound on the time-averaged regret that depends on the speed of change of the global parameter. 
\item We simulate our algorithms on a synthetic dynamic pricing data set and show that they beat other state-of-art MAB algorithms. 
\end{itemize}

\section{Related Work} \label{sec:relatedwork}
There is a wide strand of literature on MABs including finite armed stochastic MAB \cite{lairobbinsl,A2002,Auer_2002a,KL_divergence}, Bayesian MAB \cite{Kauffman,Thompson,GoyalThompson,KauffmanThompson,thompson_priorfree},
contextual MAB \cite{epoch_greedy,Slivkins,thompson_contextual} and distributed MAB \cite{tekin2013distributed, xu2015distributed, qinq_distributed}. Depending on the extent of informativeness of the arms, MABs can be categorized into three: non-informative, group informative and globally informative MABs. 

\subsection{Non-informative MAB}
We call a MAB {\em non-informative} if the reward observations of any arm do not reveal any information about the rewards of the other arms. Example of non-informative MABs include finite armed stochastic \cite{lairobbinsl,A2002} and non-stochastic \cite{auer1995gambling} MABs. Lower bounds derived for these settings point out to the impossibility of bounded regret.

\subsection{Group-informative MAB}
We call a MAB {\em group-informative} if the reward observations from an arm provides information about a group of other arms. Examples include linear contextual bandits~\cite{li2010contextual, chu2011contextual}, multi-dimensional linear bandits~\cite{abbasi2011improved, Tsiklis, dani2008stochastic, abbasi2012online, cesa2011optimal} and combinatorial bandits~\cite{chen2013combinatorial, combinatorial}. In these works, the regret is sublinear in time and in the number of arms. For example, \cite{abbasi2011improved} assumes a reward structure that is linear in an unknown parameter and shows a regret bound that scales linearly with the dimension of the parameter. It is not possible to achieve bounded regret in any of the above settings since multiple arms are required to be selected at least logarithmically many times in order to learn the unknown parameters. 

Another related work \cite{mannor2011bandits} studies a setting that interpolates between the bandit (partial feedback) and experts (full feedback) settings. In this setting, the decision-maker obtains not only the reward of the selected arm but also an unbiased estimate of the rewards of a subset of the other arms, where this subset is determined by a graph. This is not possible in our setting due to the non-linear reward structure and bandit feedback.

\subsection{Globally-informative MAB}
We call a MAB problem {\em globally-informative} if the reward observations from an arm provide information about the rewards of \textit{all} the arms \cite{lattimore2014bounded, Tsiklis_structured}. GB belongs to the class of globally-informative MAB and includes the linearly-parametrized MAB \cite{Tsiklis_structured} as a subclass. Hence, our results reduce to the results of \cite{Tsiklis_structured} for the special case when expected arm rewards are linear in the parameter. 

A related work that falls into this setting is \cite{russo2014information}, in which the authors prove regret bounds that depend on the learner's uncertainty about the optimal arm. This uncertainty depends on the learner's prior knowledge and prior observations, and affect the constant factors that contribute to the $\mathcal{O}(\sqrt{T})$ regret bound. Whereas, in our problem formulation, we show that the strong dependence of the arms through a global parameter results in bounded parameter dependent and a sub-linear worst-case regrets.

\begin{table*}[t]
\centering
{\renewcommand{\arraystretch}{0.6}
{\fontsize{10}{8}\selectfont
\setlength{\tabcolsep}{.1em}
\begin{tabular}{|l|c|c|c|c|c|}
\hline
& GB (our work) & \cite{abbasi2011improved, Tsiklis, dani2008stochastic, abbasi2012online, cesa2011optimal} & \cite{Tsiklis_structured} & \cite{paramtric_GLM} \\ \hline 
Parameter dimension & Single & Multi & Single & Multi &\\ \hline
Reward functions & Non-linear & Linear & Linear & Generalized linear & \\ \hline
Worst-case regret & BUW: $\tilde{\mathcal{O}}(\sqrt{T})$, WAGP: $\mathcal{O}(T^{1 - \frac{\gamma}{2}})$ & $\tilde{\mathcal{O}}(\sqrt{T})$ & $\mathcal{O}(\sqrt{T})$ & $\tilde{\mathcal{O}}(\sqrt{T})$\\ \hline
Parameter dependent regret & BUW: $\mathcal{O}(1)$, WAGP: $\mathcal{O}(1)$ & $\mathcal{O}\left(\log T \right)$ & $\mathcal{O}(1)$ & $\mathcal{O}\left( (\log T)^3 \right)$\\ \hline
\end{tabular}}}
\caption{Comparison with related works. $\gamma \leq 1$ represents the informativeness, which is given in Definition \ref{defn:informativeness}.}
\end{table*}

Table 1 summarizes our model and theoretical results, and compares them with the existing literature in the parametric MAB models. Although GB is more general than the model in \cite{Tsiklis_structured}, both WAGP and BUW achieves bounded parameter-dependent regret, and BUW is able to achieve the same worst-case regret as the policy in \cite{Tsiklis_structured}. On the other hand, although the linear MAB models are more general than GB, it is not possible to achieve bounded regret in these models.

\section{Problem Formulation}\label{sec:probdesc}

\subsection{Arms, Reward Functions and Informativeness}

There are $K$ arms indexed by the set $\mathcal{K} := \{1,\ldots,K\}$.
The global parameter is denoted by $\theta_{*}$, which belongs to the parameter set $\Theta$ that is taken to be the unit interval for simplicity of exposition. 
The random variable $X_{k,t}$ denotes the reward of arm $k$ at time $t$.
$X_{k,t}$ is drawn independently from a distribution $\nu_{k}(\theta_{*})$ with support $\mathcal{X}_k \subseteq [0,1]$.
The expected reward of arm $k$ is a H\"{o}lder continuous, invertible function of $\theta_{*}$, which is given by $\mu_{k}(\theta_{*}) := \mathrm{E}_{\nu_{k}(\theta_{*})}[X_{k,t}]$, where $\mathrm{E}_{\nu}[\cdot]$ denotes the expectation taken with respect to distribution $\nu$. This is formalized in the following assumption.

%We assume that $\theta_{\text{true}}$ is sampled from an unknown
%prior distribution $f(\theta)$ on $\Theta$ and we make the following
%assumptions on the reward functions $\mu_{k}$.

\begin{assumption} \label{ass:holder}
(i) For each $k \in {\cal K}$ and $\theta,\theta '\in \Theta$ there exists $D_{1,k} > 0$ and $1 < \gamma_{1,k}$, such that
$$
|\mu_{k}(\theta)-\mu_{k}(\theta')|\geq D_{1,k}|\theta-\theta'|^{\gamma_{1,k}}.
$$
(ii) For each $k\in{\cal K}$ and $\theta,\theta '\in \Theta$ there exists $D_{2,k}>0$ and $0<\gamma_{2,k}\leq1$, such that 
$$
|\mu_{k}(\theta)-\mu_{k}(\theta')|\leq D_{2,k}|\theta-\theta'|^{\gamma_{2,k}}.
$$
\end{assumption}

The first assumption ensures that the reward functions are monotonic and the second assumption, which is also known as Hölder continuity, ensures that the reward functions are smooth. These assumptions imply that the reward functions are invertible and the inverse reward functions are also Hölder continuous. Moreover, they generalize the model proposed in \cite{Tsiklis_structured}, and allow us to model real-world scenarios described in Examples 1 and 2, and propose algorithms that achieve bounded regret.

Some examples of the reward functions that satisfy Assumption \ref{ass:holder} are: (i) exponential functions such as $\mu_{k}(\theta)=a\exp(b\theta)$ where $a>0$, (ii) linear and piecewise linear functions, and (iii) sub-linear and super-linear functions in $\theta$ which are invertible in $\Theta$ such as $\mu_{k}(\theta)=a\theta^{\gamma}$ where $\gamma>0$ and $\Theta=[0,1]$.

\begin{proposition} \label{prop:func} Define $\underline{\mu}_k = \min_{\theta \in \Theta} \mu_k(\theta)$ and $\overline{\mu}_k = \max_{\theta \in \Theta} \mu_k(\theta)$. Under Assumption \ref{ass:holder}, the following are true: 
(i)  For all $k \in {\cal K}$, $\mu_k(\cdot)$ is invertible.
(ii) For all $k \in {\cal K}$ and for all $x, x' \in [\underline{\mu}_k, \overline{\mu}_k]$, 
$$
|\mu_{k}^{-1}(x)-\mu_{k}^{-1}(x')|\leq \bar{D}_{1,k}|x-x'|^{\bar{\gamma}_{1,k}}
$$
where $\bar{\gamma}_{1,k} = \frac{1}{\gamma_{1,k}}$ and $\bar{D}_{1,k} = \left(\frac{1}{D_{1,k}}\right)^{\frac{1}{\gamma_{1,k}}}$.
\end{proposition}

Invertibility of the reward functions allows us to use the rewards obtained from an arm to estimate the expected rewards of other arms. Let $\bar{\gamma}_1$ and $\gamma_2$ be the minimum exponents and $\bar{D}_1$, $D_2$ be the maximum constants, that is 
\begin{eqnarray*}
\bar{\gamma}_1 &=& \min_{k \in \mathcal{K}} \bar{\gamma}_{1,k}, \; \gamma_2 = \min_{k \in \mathcal{K}} \gamma_{2,k}, \\ 
\bar{D}_1 &=& \max_{k \in \mathcal{K}} \bar{D}_{1,k}, \; D_2 = \max_{k \in \mathcal{K}} D_{2,k}. 
\end{eqnarray*}

\begin{definition}\label{defn:informativeness} The informativeness of arm $k$ is defined as $\gamma_k := \bar{\gamma}_{1,k} \gamma_{2,k}$. The informativeness of the GB instance is defined as $\gamma := \bar{\gamma}_1 \gamma_2$.
\end{definition} 

The informativeness of arm $k$ measures the extent of information that can be obtained about the expected rewards of other arms from the rewards observed from arm $k$. As we will show later, when the informativeness is high, one can form better estimates of the expected rewards of other arms by using the rewards observed from arm $k$.

\subsection{Definition of the Regret}

The learner knows $\mu_k(\cdot)$ for all $k \in {\cal K}$ but does not know $\theta_{*}$. At each time $t$ it selects one of the arms, denoted by $I_t$, and receives the random reward $X_{I_t,t}$. 
The learner's goal is to maximize its cumulative reward up to any time $T$.

Let $\mu^{*}(\theta) := \max_{k \in \mathcal{K}} \mu_k(\theta)$ be the maximum expected reward and $\mathcal{K}^{*}(\theta) := \{k \in \mathcal{K}: \mu_k(\theta) = \mu^{*}(\theta)\}$ be the optimal set of arms for parameter $\theta$.
In addition, let $k^*(\theta)$ denote an arm that is optimal for parameter $\theta$.
We refer to the policy that selects one of the arms in $\mathcal{K}^{*}(\theta_{*})$ as the \textit{oracle} policy. The learner incurs a regret (loss) at each time it deviates from the oracle policy. We define the one-step regret at time $t$ as the difference between the expected reward of the oracle policy and the learner, which is given by $r_{t}(\theta_{*})  := \mu^{*}(\theta_{*})-\mu_{I_{t}}(\theta_{*})$. 

Based on this, the cumulative regret of the learner by time $T$ (also referred to as the regret hereafter) is defined as
\begin{align}
\text{Reg}(\theta_{*},T) :=\mathbb{E}\left[\sum_{t=1}^{T}r_{t}(\theta_{*})\right] . \notag
\end{align}
Maximizing the reward is equivalent to minimizing the regret. In the seminal work by Lai and Robbins \cite{lai1978adaptive}, it is shown that the regret becomes infinite as $T$ grows for the classical $K$-armed bandit problem. On the other hand, $\lim_{T \rightarrow \infty } \text{Reg}(\theta_{*},T) < \infty$ will imply that the learner deviates from the oracle policy only finitely many times. In the following sections, we prove that this holds for GB.

\section{Weighted-Arm Greedy Policy (WAGP)} \label{sec:WAGP}

\begin{algorithm}[t]
\caption{The WAGP}
	\label{fig:GP}
	\normalsize
	\begin{algorithmic}[1]
		\STATE {\bfseries Inputs:} $\mu_k(\cdot)$ for each arm $k$
		\STATE {\bfseries Initialization:} $w_k(0)=0,\hat{\theta}_{k,0}=0, \hat{X}_{k,0}=0 ,N_k(0)=0$ for all $k\in{\cal K}$, $t=1$
		\WHILE{$t>0$}
		\IF{$t = 1$}
			\STATE Select arm $I_1$ uniformly at random from ${\cal K}$
		\ELSE
			\STATE Select arm $I_{t} \in \argmax_{k\in{\cal K}} \mu_{k}(\hat{\theta}_{t-1})$ (break ties randomly)
		\ENDIF
		\STATE $\hat{X}_{k,t}=\hat{X}_{k,t-1}$ for all $k \in {\cal K}\setminus I_t$
		\STATE $\hat{X}_{I_t,t} = \frac{N_{I_t}(t-1) \hat{X}_{I_t,t-1} + X_{I_{t},t}}{N_{I_t}(t-1) +1}$
		\STATE $\hat{\theta}_{k,t} = \arg\min_{\theta \in \Theta} |\mu_k(\theta) - \hat{X}_{k,t}|$ for all $k\in{\cal K}$
		\STATE $N_{I_{t}}(t)=N_{I_{t}}(t-1)+1$ 
		\STATE $N_{k}(t)=N_{k}(t-1)$ for all $k \in {\cal K} \setminus I_{t}$
		\STATE $w_{k}(t)=N_{k}(t)/t$ for all $k\in{\cal K}$
		\STATE $\hat{\theta}_{t}=\sum_{k=1}^{K}w_{k}(t)\hat{\theta}_{k,t}$
		\ENDWHILE
	\end{algorithmic}
\end{algorithm}

In this section, we propose a greedy policy called the Weighted-Arm Greedy Policy (WAGP). The pseudocode of WAGP is given in Algorithm \ref{fig:GP}. The WAGP consists of two phases: arm selection phase and parameter update phase. 

Let $N_k(t)$ denote the number of times arm $k$ is selected until time $t$, and $\hat{X}_{k,t}$ denote the reward estimate, $\hat{\theta}_{k,t}$ denote the global parameter estimate and $w_k(t)$ denote the weight of arm $k$ at time $t$. Initially, all the counters and estimates are set to zero. In the arm selection phase at time $t > 1$, the WAGP selects the arm with the highest estimated expected reward:  
$
I_t \in \argmax_{k \in {\cal K}} \mu_{k}(\hat{\theta}_{t-1})
$
where $\hat{\theta}_{t-1}$ is the estimate of the global parameter calculated at the end of time $t-1$.\footnote{The ties are broken randomly.}$^{,}$\footnote{For $t=1$, the WAGP selects a random arm since there is no prior reward observation that can be used to estimate $\theta_{*}$.}

In the parameter update phase the WAGP updates: (i) the estimated reward of selected arm $I_t$, denoted by $\hat{X}_{I_t, t}$, (ii) the global parameter estimate of the selected arm $I_t$, denoted by $\hat{\theta}_{I_t,t}$, (iii) the global parameter estimate $\hat{\theta}_t$, and (iv) the counters $N_k(t)$. The reward of estimate of arm $I_t$ is updated as: 
$$
\hat{X}_{I_t, t} = \frac{N_{I_t}(t-1) \hat{X}_{I_t, t-1} + X_{I_t,t}}{N_{I_t}(t-1) +1}. 
$$
The reward estimates of the other arms are not updated. The WAGP constructs estimates of the global parameter from the rewards of all the arms and combines their estimates using a weighted sum. The WAGP updates $\hat{\theta}_{I_t,t}$ of arm $I_t$ in a way that minimizes the distance between $\hat{X}_{I_t,t}$ and $\mu_{I_t}(\theta)$, i.e.,  
$
\hat{\theta}_{I_t,t} = \arg\min_{\theta \in \Theta} |\mu_{I_t}(\theta) - \hat{X}_{I_t,t}| .
$
Then, the WAGP sets the global parameter estimate as
$
\hat{\theta}_{t}=\sum_{k=1}^{K}w_{k}(t)\hat{\theta}_{k,t}
$
where $w_k(t) = N_k(t)/t$. Hence, the WAGP gives more weights to the arms with more reward observations since the confidence on their estimates are higher. 

\section{Regret Analysis of the WAGP} \label{sec:WAGPregret}

\subsection{Preliminaries for the Regret Analysis}

In this subsection we define the tools that will be used in deriving the regret bounds for the WAGP. Consider any arm $k\in{\cal K}$. Its \emph{optimality region} is defined as 
\begin{align}
\Theta_{k}:=\{\theta\in\Theta : k \in \mathcal{K}^{*}(\theta)\}. \notag
\end{align}
Note that $\Theta_{k}$ can be written as union of intervals in each of which arm $k$ is optimal. Each such interval is called an {\em optimality interval}. Clearly, we have $\bigcup_{k\in{\cal K}}\Theta_{k}=\Theta$. If $\Theta_{k}=\emptyset$ for an arm $k$, this implies that there exists no global parameter
value for which arm $k$ is optimal. Since there exists an arm $k'$ such that $\mu_{k'}(\theta) >\mu_k(\theta)$ for any $\theta \in \Theta$ for an arm with $\Theta_k = \emptyset$, the greedy policy will discard arm $k$ after $t= 1$. Therefore, without loss of generality we assume that $\Theta_{k}\neq\emptyset$
for all $k\in{\cal K}$. 
The \emph{suboptimality gap} of arm $k\in{\cal K}$ given global parameter $\theta_{*} \in \Theta$ is defined as $\delta_{k}(\theta_{*}):=\mu^{*}(\theta_{*})-\mu_{k}(\theta_{*})$.
The {\em minimum suboptimality gap} given global parameter $\theta_{*} \in \Theta$ is defined as
$\delta_{\text{min}}(\theta_{*}):=\min_{k\in{\cal K}\setminus \mathcal{K}^{*}(\theta_{*})}\delta_{k}(\theta_{*})$.

\begin{figure}[h]
\begin{centering}
 \includegraphics[clip,width=1\columnwidth,trim = 50mm 20mm 30mm 30mm]{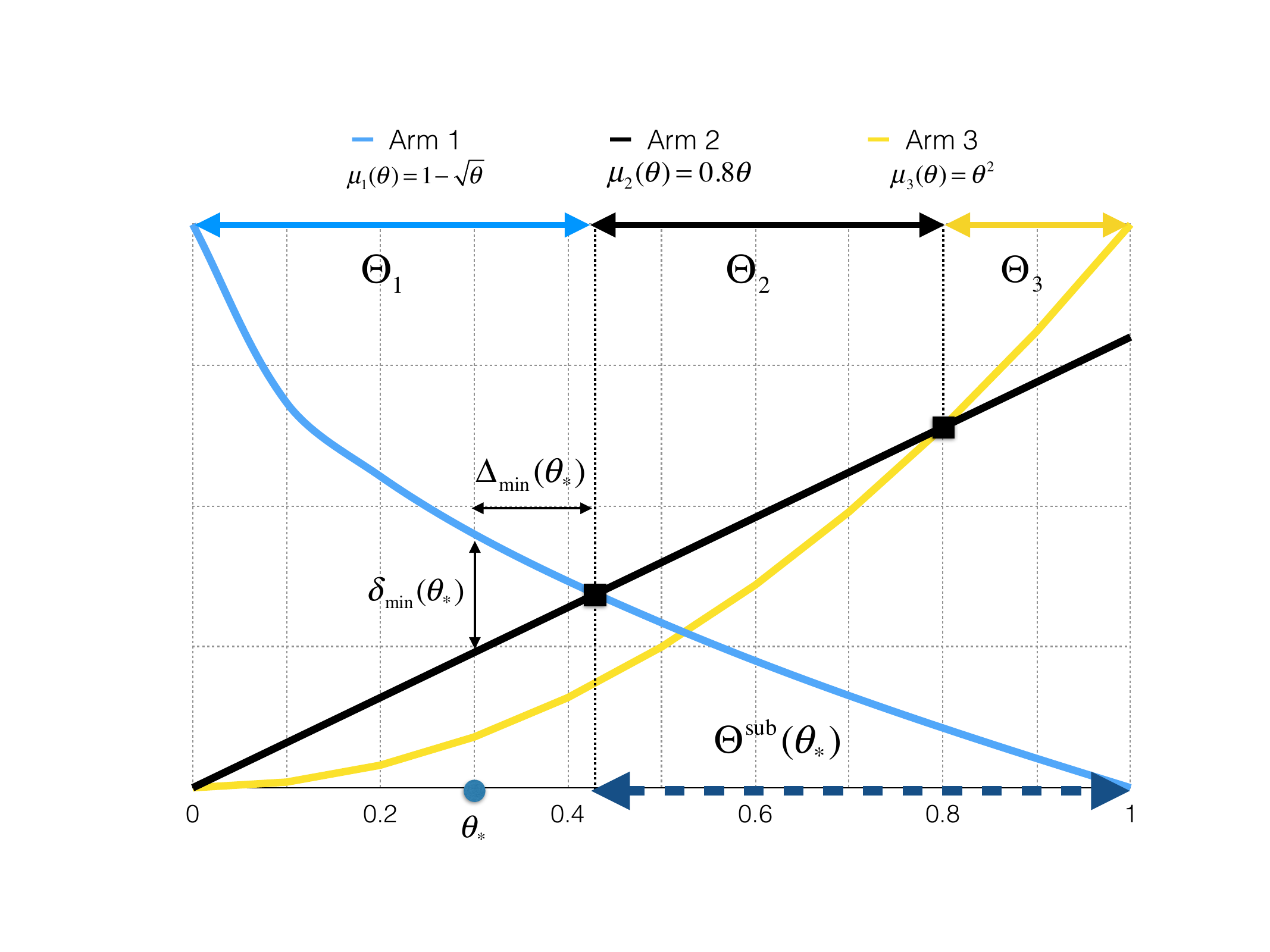}
 \protect\caption{Illustration of the minimum suboptimality gap and the suboptimality distance.}
\label{fig: illustration} 
\end{centering}
\end{figure}

Let $\Theta^{\text{sub}}(\theta_{*})$ be the suboptimality region of the global parameter $\theta_{*}$, which is defined as the subset of the parameter space in which none of the arms in $\mathcal{K}^{*}(\theta_{*})$ is optimal, i.e.,
\begin{align}
\Theta^{\text{sub}}(\theta_{*}) := \Theta \setminus \bigcup_{k' \in \mathcal{K}^{*}(\theta_{*})} \Theta_{k'}. \notag
\end{align}
We will show that as time proceeds, the global parameter estimate will converge to $\theta_{*}$. However, if $\theta_{*}$ lies close to $\Theta^{\text{sub}}(\theta_{*})$, the global parameter estimate may fall into the suboptimality region for a large number of times, thereby resulting in a large regret.
In order to bound the expected number of times this happens, we define the \emph{suboptimality distance} as the smallest distance between the global parameter and the suboptimality region.

\begin{definition} \label{def:dist} For a given global parameter
$\theta_{*}$, the \textit{suboptimality distance} is defined as 
\[
\Delta_{\text{min}}(\theta_{*}):=\left\{ \begin{array}{lr}
\inf_{\theta'\in\Theta^{\text{sub}}(\theta_{*})} |\theta_{*}-\theta'| & \text{if } \Theta^{\text{sub}}(\theta_{*}) \neq \emptyset\\
1 & \text{if }\Theta^{\text{sub}}(\theta_{*}) = \emptyset
\end{array}\right.
\]
\end{definition}

From the definition of the suboptimality distance it is evident that the proposed policy always selects an optimal arm in ${\cal K}^{*}(\theta_{*})$ when $\hat{\theta}_{t}$ is within $\Delta_{\text{min}}(\theta_{*})$ of $\theta_{*}$. For notational brevity, we also use $\Delta_{*} :=\Delta_{\text{min}}(\theta_{*})$ and $\delta_{*} := \delta_{\text{min}}(\theta_{*})$. An illustration of the suboptimality gap and the suboptimality distance is given in Fig. \ref{fig: illustration} for the case with $3$ arms and reward functions $\mu_{1}(\theta)=1-\sqrt{\theta}$, $\mu_{2}(\theta)=0.8\theta$ and $\mu_{3}(\theta)=\theta^{2}$, $\theta \in [0,1]$. 

The notations frequently used in the regret analysis is highlighted in Table 2. 

\begin{table}[h]
\centering
{\renewcommand{\arraystretch}{0.8}
{\fontsize{10}{8}\selectfont
\setlength{\tabcolsep}{.1em}
\begin{tabular}{|l|c|c|}
\hline
$\mathcal{K}^{*}(\theta_{*})$ & set of optimal arms for $\theta_{*}$ \\ \hline
$\mu^{*}(\theta_{*})$ & expected reward of optimal arms \\ \hline
$I_t$ & selected arm at time $t$ \\ \hline
$\hat{\theta}_t$ & global parameter estimate at time $t$ \\ \hline
$\delta_{*} = \delta_{\text{min}}(\theta_{*})$ & minimum suboptimality gap \\ \hline
$\Delta_{*} = \Delta_{\text{min}}(\theta_{*})$ & minimum suboptimality distance \\ \hline
$\Theta_k$ & optimality region of arm $k$ \\ \hline
$\Theta^{\text{sub}}(\theta_{*})$ & suboptimality region of $\theta_{*}$ \\ \hline
$\gamma$ & informativeness of the arms \\ \hline
\end{tabular}}}
\caption{Frequently used notations in regret analysis}
\end{table}

\subsection{Worst-case Regret Bounds for the WAGP} \label{sub:parameter_free}

First, we show that parameter estimate of the WAGP converges in the mean-squared sense. 

\begin{theorem} \label{thm:mean_squared}
Under Assumption \ref{ass:holder}, the global parameter estimate of the WAGP converges to true value of global parameter in mean-squared sense, i.e., $
\lim_{t \rightarrow \infty} \mathbb{E}\left[|\hat{\theta}_t - \theta_{*}|^2\right] = 0.
$
\end{theorem}

The following theorem bounds the expected one-step regret of the WAGP.

\begin{theorem} Under Assumption \ref{ass:holder}, we have for WAGP $\mathbb{E}\left[r_{t}(\theta_{*})\right]\leq \mathcal{O}(t^{-\frac{\gamma}{2}})$. 
 \label{thm:onestepregret} 
\end{theorem}

Theorem \ref{thm:onestepregret} proves that the expected one-step regret of the WAGP converges to zero.\footnote{The asymptotic notation is only used for a succinct representation, to hide the constants and highlight the time dependence. This bound holds not just asymptotically but for any finite $t$.} This is a \emph{worst-case} bound in the sense that it holds for any $\theta_{*}$. Using this result, we derive the following worst-case regret bound for the WAGP.

\begin{theorem} \label{thm:par_indep} Under Assumption \ref{ass:holder}, the worst-case
regret of WAGP is 
\begin{align}
\sup_{\theta_{*} \in \Theta} \text{Reg}(\theta_{*},T) \leq \mathcal{O}(K^{\frac{\gamma}{2}}T^{1-\frac{\gamma}{2}}) .     \notag
\end{align}
\end{theorem}
Note that the worst-case regret bound is sublinear both in the time horizon $T$ and the number of arms $K$. Moreover, it depends on the  informativeness $\gamma$. When the reward functions are linear or piecewise linear, we have $\gamma=1$, which is an extreme case of our model; hence, the worst-case regret is $\mathcal{O}(\sqrt{T})$, which matches with (i) the worst-case regret bound of the standard MAB algorithms in which a linear estimator is used \cite{BubeckBianchi}, and (ii) the bounds obtained for the linearly parametrized bandits \cite{Tsiklis_structured}. 

\subsection{Parameter Dependent Regret Bounds for the WAGP} \label{subsec:par_dep}

In this section we bound the parameter dependent regret of the WAGP. 
First, we introduce several constants that will appear in the regret bound. 
\begin{definition} $C_{1}(\Delta_{*})$ is
the smallest integer $\tau$ such that $\tau\geq \left( \frac{\bar{D}_1 K}{\Delta_{*}}\right)^{\frac{2}{\bar{\gamma}_1}} \frac{\log(\tau)}{2}$
and $C_{2}(\Delta_{*})$ is the smallest integer $\tau$ such that
$\tau\geq \left( \frac{\bar{D}_1 K}{\Delta_{*}}\right)^{\frac{2}{\bar{\gamma}_1}} \log(\tau)$.
\end{definition}

Closed form expressions for these constants can be obtained in terms of the \textit{glog} function \cite{kalman2001generalized}, for which the following equivalence holds: $y = \operatorname{glog}(x)$ if and only if $x = \frac{\exp(y)}{y}$. Then, we have 
\begin{align}
& C_1(\Delta_{*}) = \Big\lceil \frac{1}{2} \left( \frac{\bar{D}_1 K}{\Delta_{*}}\right)^{\frac{2}{\bar{\gamma}_1}} \operatorname{glog}\left(\frac{1}{2} \left( \frac{\bar{D}_1 K}{\Delta_{*}}\right)^{\frac{2}{\bar{\gamma}_1}}\right) \Big\rceil, \notag \\
& C_2(\Delta_{*}) = \Big\lceil \left( \frac{\bar{D}_1 K}{\Delta_{*}}\right)^{\frac{2}{\bar{\gamma}_1}} \operatorname{glog}\left(\left( \frac{\bar{D}_1 K}{\Delta_{*}}\right)^{\frac{2}{\bar{\gamma}_1}}\right) \Big\rceil,. \notag
\end{align}

Next, we define the expected regret incurred between time steps $T_{1}$ and $T_{2}$ given $\theta_{*}$ as 
$
R_{\theta_{*}}(T_{1},T_{2}):= \sum_{t=T_1}^{T_2} \mathbb{E}\left[r_t(\theta_{*})\right]. \notag
$
The following theorem bounds the parameter dependent regret of the WAGP.
\begin{theorem} \label{thm:par_dep} Under Assumption \ref{ass:holder}, the regret
of the WAGP is bounded as follows: \\
(i) For $1\leq T < C_{1}(\Delta_{*})$, the regret grows sublinearly in
time, i.e., 
\begin{align}
R_{\theta_{*}}(1,T) \leq S_1 + S_2 T^{1 - \frac{\gamma}{2}}  \notag
\end{align}
where $S_1$ and $S_2$ are constants that are independent of the global parameter $\theta_{*}$, whose exact forms are given in Appendix \ref{app:theorem4proof}.

(ii) For $C_{1}(\Delta_{*})\leq T <  C_{2}(\Delta_{*})$, the regret grows logarithmically in time, i.e., 
\begin{align}
R_{\theta_{*}}(C_{1}(\Delta_{*}), T)\leq1+2K\log \left(\frac{T}{C_{1}(\Delta_{*})} \right) . \notag
\end{align}

(iii) For $T \geq C_{2}(\Delta_{*})$, the growth of the regret is bounded, i.e., 
\begin{align}
R_{\theta_{*}}(C_{2}(\Delta_{*}), T)\leq K\frac{\pi^{2}}{3}. \notag
\end{align} 

Thus, we have $\lim_{T\rightarrow\infty}\text{Reg}(\theta_{*},T)<\infty$, i.e., $\text{Reg}(\theta_{*},T) = \mathcal{O}(1)$.
\end{theorem}

Theorem \ref{thm:par_dep} shows that the regret is inversely proportional to the suboptimality distance $\Delta_{*}$, which depends on $\theta_{*}$. The regret bound contains three regimes of growth: Initially the regret grows sublinearly until time threshold $C_{1}(\Delta_{*})$. After this, it grows logarithmically until time threshold $C_2(\Delta_{*})$. Finally, the growth of the regret is bounded after time threshold $C_2(\Delta_{*})$. In addition, since $\lim_{\Delta_{*}\rightarrow0}C_{1}(\Delta_{*})=\infty$, in the worst-case, the bound given in Theorem \ref{thm:par_dep} reduces to the one given in Theorem \ref{thm:par_indep}. 
It is also possible to calculate a Bayesian risk bound for the WAGP by assuming a prior over the global parameter space. This risk bound is given to be $\mathcal{O}(\log T)$ when $\gamma = 1$ and $\mathcal{O}(T^{1 - \gamma})$ when $\gamma < 1$ (see \cite{atan2015}).

\begin{theorem} \label{thm:convergence} The sequence of arms selected by the WAGP converges to the optimal arm almost surely, i.e., $\lim_{t\rightarrow\infty}I_t \in \mathcal{K}^{*}(\theta_{*})$ with probability 1. 
\end{theorem}

Theorem \ref{thm:convergence} implies that a suboptimal arm is selected by the WAGP only finitely many times. This is the major difference between GB and the classical MAB \cite{lairobbinsl,A2002,russo2014information}, in which every arm needs to be selected infinitely many times asymptotically by any {\em good} learning algorithm.

\begin{remark} Assumption \ref{ass:holder} ensures that the parameter dependent regret is bounded. When this assumption is relaxed, bounded regret may not be achieved, and the best possible regret becomes logarithmic in time. For instance, consider the case when the reward functions are constant over the global parameter space, i.e., $\mu_{k}(\theta_{*})=m_{k}$ for all $\theta_{*} \in [0,1]$ where $m_k$ is a constant. This makes the reward functions non-invertible. In this case, the learner cannot use the rewards obtained from the other arms when estimating the rewards of arm $k$. Thus, it needs to learn $m_k$ of each arm separately, which results in logarithmic in time regret when a policy like UCB1 \cite{A2002} is used. This issue still exists even when there are only finitely many possible solutions to $\mu_{k}(\theta_{*}) = x$ for some $x$, in which case some of the arms should be selected at least logarithmically many times to rule out the incorrect global parameters.
\end{remark}

\subsection{Lower Bound on the Worst-case Regret}

Theorem \ref{thm:par_indep} shows that the worst-case regret of the WAGP is $\mathcal{O}(T^{1- \frac{\gamma}{2}})$, which implies that the regret decreases with $\gamma$. In this section, we give lower bounds on the parameter dependent and the worst-case regrets.

\begin{theorem} \label{thm:lower}
For $T \geq 8$ and any policy, the parameter dependent regret is lower bounded by $\Omega(1)$ and the worst-case regret is lower bounded by $\Omega(\sqrt{T})$.
\end{theorem}
The theorem above raises a natural question: Can we achieve both $\tilde{\mathcal{O}}(\sqrt{T})$ worst-case regret (like the UCB based MAB algorithms \cite{A2002}) and bounded parameter dependent regret by using a combination of UCB and WAGP policies? We answer this question in the affirmative in the next section.

\section{The Best of the UCB and the WAGP (BUW)} \label{sec:BUW}
In the this section, we propose the Best of the UCB and the WAGP (BUW), which combines UCB1 and the WAGP to achieve bounded parameter dependent and $\mathcal{O}(\sqrt{T})$ worst-case regrets. In the worst-case, the WAGP achieves $\mathcal{O}(T^{1 - \frac{\gamma}{2}})$ regret, which is weaker than $\tilde{\mathcal{O}}(\sqrt{T})$ worst-case regret of UCB1. On the other hand, the WAGP achieves bounded parameter dependent regret whereas UCB1 achieves a logarithmic parameter dependent regret. In this section, we propose an algorithm which combines these two algorithms and achieves both $\tilde{\mathcal{O}}(\sqrt{T})$ worst-case regret and bounded parameter dependent regret. 

The main idea for such an algorithm follows from Theorem \ref{thm:par_dep}. Recall that Theorem \ref{thm:par_dep} shows that the WAGP achieves $O(T^{1 - \frac{\gamma}{2}})$ regret when $1 <T < C_1(\Delta_{*})$. If the BUW could follow the recommendations of UCB1 when $T < C_1(\Delta_{*})$ and the recommendations of the WAGP when $T \geq C_1(\Delta_{*})$, then it will achieve a worst-case regret bound of $\tilde{\mathcal{O}}(\sqrt{T})$ and bounded parameter-dependent regret. The problem with this approach is that the suboptimality distance $\Delta_{*}$ is unknown a priori. We can solve this problem by using a data-dependent estimate $\tilde{\Delta}_t$ where $\Delta_{*} > \tilde{\Delta}_t$ holds with high probability. The data-dependent estimate $\tilde{\Delta}_t$ is given as
$$
\tilde{\Delta}_t = \hat{\Delta}_t - \bar{D}_1 K \left(\frac{\log t}{t}\right)^{\frac{\bar{\gamma}_1}{2}}
$$
where 
\[
 \hat{\Delta}_t = \Delta_{\text{min}}(\hat{\theta}_t)=\left\{ \begin{array}{lr}
\inf_{\theta'\in\Theta^{\text{sub}}(\hat{\theta}_t)} |\hat{\theta}_t-\theta'| & \text{if } \Theta^{\text{sub}}(\hat{\theta}_t) \neq \emptyset\\
1 & \text{if }\Theta^{\text{sub}}(\hat{\theta}_t) = \emptyset
\end{array}\right.
\]

\begin{algorithm}[t]
\caption{The BUW}
	\label{fig:BUW}
	\normalsize
	\begin{algorithmic}[1]
	       \STATEx {\bfseries Inputs: } $T$, $\mu_k(\cdot)$ for each arm $k$. 
	       \STATEx {\bfseries Initialization: } Select each arm once for $t = 1,2,\ldots, K$, compute $\hat{\theta}_{k,K}$, $N_k(K)$, $\hat{\mu}_k$, $\hat{X}_{k,K}$ for all $k \in {\cal K}$, and $\hat{\theta}_K$, $\hat{\Delta}_K$, $\tilde{\Delta}_K$, $t=K+1$
	       \WHILE{ $t \geq K+1$}
	        \IF{$t < C_2\left(\max\left( 0, \tilde{\Delta}_{t-1}\right)\right)$}
	        \STATE $I_t \in \arg\max_{k \in \mathcal{K}} \hat{X}_{k,t-1} + \sqrt{\frac{2 \log (t-1)}{N_k(t-1)}}$
		\ELSE
		\STATE $I_t \in \arg\max_{k \in \mathcal{K}} \mu_k(\hat{\theta}_{t-1})$
		\ENDIF
		\STATE Update $\hat{X}_{I_t,t}$, $N_k(t)$, $w_k(t)$, $\hat{\theta}_{k,t}$, $\hat{\theta}_t$ as in the WAGP
		\STATE Solve
		\[
	 		\hat{\Delta}_t =\left\{ \begin{array}{lr}
			\inf_{\theta'\in\Theta^{\text{sub}}(\hat{\theta}_t)} |\hat{\theta}_t-\theta'| & \text{if } \Theta^{\text{sub}}(\hat{\theta}_t) \neq \emptyset\\
			1 & \text{if }\Theta^{\text{sub}}(\hat{\theta}_t) = \emptyset
			\end{array}\right.
		\]
		\STATE $\tilde{\Delta}_t = \hat{\Delta}_t - \bar{D}_1 K \left( \frac{ \log t}{t}\right)^{\frac{\bar{\gamma}_1}{2}}$       
		\ENDWHILE
	\end{algorithmic}
\end{algorithm}

The pseudo-code for the BUW is given in Fig. \ref{fig:BUW}. The regret bounds for the BUW are given in the following theorem. 
%In theorem, we drop the dependence on the $\theta_{*}$ in order to write the regret bounds more elegantly.

\begin{theorem} \label{thm:BUW}
Under Assumption 1, the worst-case regret of the BUW is bounded as follows: 
$$
\sup_{\theta_{*} \in \Theta} \text{Reg}(\theta_{*},T) \leq \tilde{\mathcal{O}}(\sqrt{KT}).
$$
Under Assumption 1, the parameter dependent regret of the BUW is bounded as follows:

(i) For $1 \leq T <  C_{2}(\Delta_{*}/3)$, the regret grows logarithmically in time, i.e., 
\begin{align}
R_{\theta_{*}}(1, T) \leq \left[8 \sum_{k: \mu_k < \mu^{*}} \frac{\log T}{\delta_k} \right] + K \left(1 +  \pi^2 \right) . \notag
\end{align}
(ii) For $T \geq C_{2}(\Delta_{*}/3)$, the growth of the regret is bounded, i.e., 
\begin{align}
R_{\theta_{*}}(C_{2}(\Delta_{*}/3), T) \leq K \pi^2 . \notag
\end{align}
\end{theorem}
The BUW achieves the lower bound given in Theorem 6, that is $\mathcal{O}(1)$ parameter-dependent regret and $\tilde{\mathcal{O}}(\sqrt{T})$ worst-case regret. 

\section{Extension: Learning under Time-varying Global Parameter} \label{sec:ext}
In this section, we consider the case when the global parameter slowly changes over time. 

\subsection{Time-varying Global Parameter}
We denote the global parameter at time $t$ as $\theta_{*}^t$. The reward of arm $k$ at time $t$, i.e., $X_{k,t}$, is drawn independently from the distribution $\nu_k(\theta_{*}^{t})$ where $\mathrm{E} [X_{k,t}] = \mu_k(\theta_{*}^{t})$. In order to bound the regret, we impose a restriction on the {\em speed} of change of the global parameter which is formalized in the following assumption. 

\begin{assumption} \label{ass:drift}
For any $t$ and $t'$, we have 
\begin{align}
|\theta_{*}^{t} - \theta_{*}^{t'}| \leq \left| \frac{t}{\tau} -\frac{t'}{\tau} \right|       \notag
\end{align}
where $\tau>0$ controls the speed of the change. 
\end{assumption}

In the static global parameter model, we were able to bound the problem specific regret with a finite constant number (independent of time horizon $T$) and the parameter-independent regret with a sublinear function of time. However, when the global parameter is changing, it is not possible to obtain these bounds. Therefore, we focus on the average regret, which is given as 
\begin{align}
\text{Reg}^{\text{ave}}(T) := \frac{1}{T} \mathbb{E}\left[\sum_{t=1}^{T} \mu^{*}(\theta_{*}^{t})-\sum_{t=1}^{T}\mu_{I_{t}}(\theta_{*}^{t}) \right] . \notag
\end{align}
The WAGP needs to be modified to handle the non-stationary global parameter since the optimal arms $\mathcal{K}^{*}(\theta_{*}^{t})$ may change over time. 

\subsection{Description and Regret of the Non-stationary WAGP}
The non-stationary WAGP uses only a recent past window of reward observations when estimating the global parameter~\cite{garivier2008upper}. By choosing the window length appropriately, we can balance the regret due to the variation of the global parameter over time given in Assumption \ref{ass:drift} and the sample size within the window. The non-stationary WAGP groups the time steps into rounds $\rho=1,2,\ldots$, each having a fixed length of $2\tau_{h}$, where $\tau_{h}$ is called \textit{half window length}. The key point in the modified algorithm is to keep separate counters for each round and estimate the global parameter in a round based only on observations that are made within the particular window of each round. Each round $\rho$ is further divided into two sub-rounds. The first sub-round is called passive sub-round, while the second one is called the active sub-round. The first round, $\rho=0$, is an exception where it is both an active and a passive sub-round. 

\begin{figure}[h]
\begin{center}
 \includegraphics[clip,width=1.2\columnwidth,trim = 50mm 130mm 30mm 30mm]{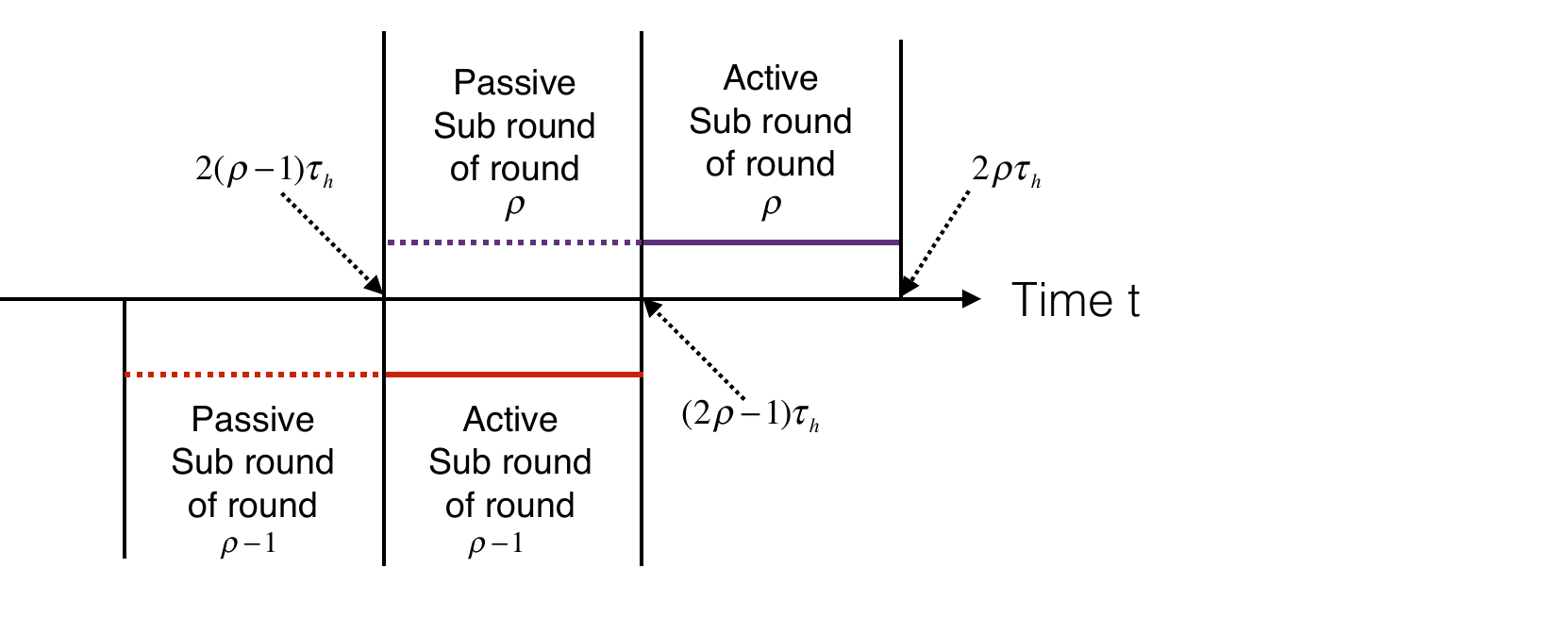}
 \protect\caption{Operation of the non-stationary WAGP.}
 \end{center}
\label{fig: rounding} 
\end{figure}

A different instance of the modified WAGP is run in each round. Let $\text{WAGP}_{\rho}$ be the running instance of the modified WAGP at round $\rho$. The arm selected at time $t$ is based on $\text{WAGP}_{\rho}$ if time $t$ is in the active sub-round of round $\rho$. Let $N_{k,\rho}(t)$ and $\hat{X}_{k,\rho,t}$ be the number of times arm $k$ is chosen and the estimate of the arm $k$ at round $\rho$ at time $t$, respectively. At the beginning of each round $\rho$, the estimates and counters of that round are set to zero, i.e., $N_{k,\rho}\left(2\tau_{h}(\rho-1)\right)=0$ and $\hat{X}_{k,\rho,2\tau_{h}(\rho-1)}=0$. However, due to the subround structure, the learner can use the observations from the passive subround of a round when choosing actions in the active subround of a round.

Similar to static parameter case, the WAGP selects the arm with the highest estimated reward. Let $\hat{\theta}_{k,\rho,t}$ denote the parameter estimate from arm $k$ at round $\rho$ at time $t$, which is given as $\arg\min_{\theta \in \Theta} |\mu_k(\theta) - \hat{X}_{k,\rho,t}|$.

 The global parameter estimate at round $\rho$ is then given by 
$
\hat{\theta}_{\rho, t} = \sum_{k=1}^K w_{k,\rho}(t) \hat{\theta}_{k,\rho,t}, \notag
$
where $w_{k,\rho}(t) = N_{k,\rho}(t)/(t - 2\tau_{h}(\rho-1))$. The arm with the highest reward estimate at round $\rho$ is selected, i.e., 
$
I_t = \arg\max_{k \in \mathcal{K}} \mu_k(\hat{\theta}_{\rho, t-1}) \notag
$
%Next theorem quantifies the average regret bound with respect to the stability and exponent of the drift. 

\begin{theorem} \label{thm:drift} Under Assumptions \ref{ass:holder} and \ref{ass:drift}, when the half window length  of the non-stationary WAGP is set to $\tau_h= \lceil \tau^{ \frac{ \gamma_2}{( \gamma_2+0.5)}} \rceil$, the average regret is
$
\text{Reg}^{\text{ave}}(T) 
\leq \mathcal{O} \left(\tau^{\frac{- \gamma \gamma_2 }{(2 \gamma_2+1)}} \right) .   \notag
$
\end{theorem}

Theorem \ref{thm:drift} shows that the average regret is bounded by a decreasing function of $\tau$ and informativeness. This is expected since the greedy policy is able to track the changes in the parameter when the drift is slow. Note that tracking performance of non-stationary WAGP depends on the informativeness because it is directly related to learning rate of the global parameter.

\section{Illustrative Results : A Dynamic Pricing Example} \label{sec:numerical}

To the best of our knowledge, there are currently no public benchmarks to test bandit algorithms on real world data. This is because the real world data does not contain the rewards of the arms that are not selected in the real time -- the counterfactuals. Hence, bandit algorithms are generally tested on synthetic datasets \cite{Tsiklis_structured, lattimore2014bounded, paramtric_GLM}. 

\subsection{Synthetic Dynamic Pricing Data}

We perform experiments on synthetic data inspired by the dynamic pricing example formulated in Section 1. We assume that the expected sales $S_{p,t}$ at time $t$ under price $p$ are of the form $\mathbb{E}\left[S_{p,t} \right] = (1- p\theta_{*})^2$, where $\theta_{*}$ characterizes the market size, and is set to $0.4$. Note that this is the linear-power demand model used by \cite{song2008structural, huang2013demand}. The expected revenue is $\mathbb{E} \left[R_{p,t} \right] = p (1- p\theta_{*})^2$. Note that reward function is $\mu_p = \mu_p(\theta_{*})= p (1- p \theta_{*})^2$ for this problem instance. We generate random rewards of each price $p$ at each time $t$ by drawing randomly from a Beta distribution with parameters $1$ and $(1-\mu_p)/\mu_p$, i.e., $R_{p,t} \sim \operatorname{Beta}(1,(1-\mu_p)/\mu_p)$, and hence $\mathbb{E}\left[ R_{p,t} \right] = \mu_p$. We set $K=12$ with $0.4,0.45,0.5, \ldots, 0.95$. 

\subsection{Results}
\textbf{Experiment 1 (Comparison)}: We compare our algorithm with two different benchmarks: UCB1 \cite{A2002} and Uncertainty Ellipsoid (UE) \cite{Tsiklis}. UCB1 treats each arm independently and learn their expected rewards by exploration. UE is proposed for linearly parametrized reward structure with high-dimensional parameter space. In our setting, UE can be used by setting an arm vector $u_p = \left[ p, p^2, p^3 \right]$ in order to fit a polynomial with order $3$ for the expected rewards. We generate rewards according to the setting described above and average the results over $100$ iterations. Fig. $4$ shows that the WAGP significantly outperforms UCB1 by exploiting the correlations between the arms. The significant performance advantage obtained by the WAGP as compared to UCB1 is due to the fact that the WAGP is able to focus on good arms early on while UCB1 learns each arm separately. The WAGP selects arm 10 (the best arm) $81.7\%$ of time, arm 9 (the second best arm) $16.4\%$ of time and the rest of the arms $1.9\%$ of time. UE outperforms UCB1 by using (some) of the correlations between the arms, however, fails to achieve the performance of the WAGP. The reason is that the WAGP learns about the parameter by selecting any of the arms, however, UE needs to select $3$ linearly independent arms in order to learn about the parameter. 

\begin{figure}[h]
\begin{center}
 \includegraphics[width=1\columnwidth]{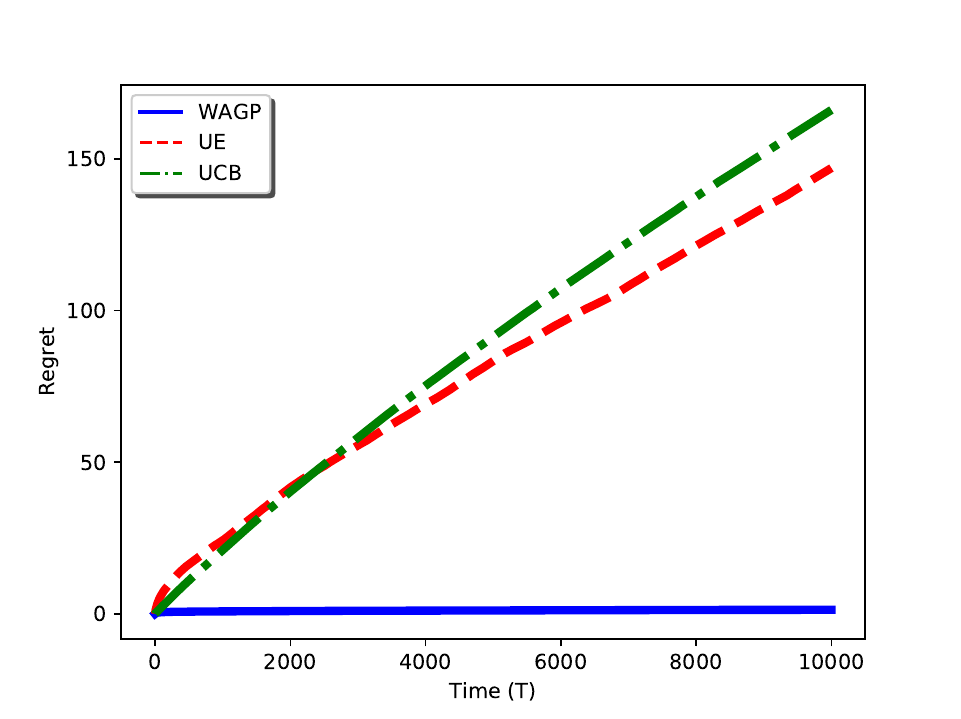}
 \protect\caption{Comparison of UCB1, UE and the WAGP for dynamic pricing example on $10000$ samples}
 \end{center}
 \label{fig:comp1} 
\end{figure}

\textbf{Experiment 2 (The effect of the suboptimality distance)}: Table $3$ shows the regret of the WAGP for different $\theta_{*}$ and hence different $\Delta_{*}$. From this, it can be seen that the regret of the WAGP is indeed decreasing with the suboptimality distance as predicted by Theorem 4.

\begin{table}[ht]
\centering{}{\fontsize{9}{8}\selectfont %
\begin{tabular}{|l|l|l|l|l|l|}
\hline
$\theta_{*}$ & $0.2$  & $0.1$  & $0.3$  & $0.8$  & $0.5$ \tabularnewline \hline 
$\Delta_{*}$  & $0.17$  & $0.1$  & $0.07$  & $0.02$  & $0.01$ \tabularnewline \hline
Regret & $0.3$ & $0.65$ & $0.72$ & $2.02$ & $2.47$ \tabularnewline \hline
\end{tabular}} 
\label{table:comp}
\vspace{0.1in}
\protect\caption{Regret of the WAGP for different $\theta_{*}$ on $10000$ samples}
\end{table} 

\textbf{Experiment 3 (Non-stationary Parameter)}: In this part, we show the performance of the proposed methods for a non-stationary setting. The expected revenue for price $p$ at time $t$ is given by $\mathbb{E}\left[ R_{p,t} \right] = p (1- p\theta_{*}^t )^2 $. We assume that $\theta^{*}_1 = 0.5$ and $\theta_{*}^{t}= \theta_{*}^{t-1} + Y_t/\tau$ where $Y_t$ is a random variable with $\Pr(Y_t =1) = 0.6$ and $\Pr(Y_t =-1) = 0.4$ and $\tau >0$. Hence, 
\begin{align}
|\theta_{*}^{t} - \theta_{*}^{t'}| \leq \left| \frac{t}{\tau} -\frac{t'}{\tau} \right| \notag
\end{align}
with probability $1$ for all $t,t'\geq 1$. 

\begin{figure*}[t]
\begin{subfigure}{.5\textwidth}
  \centering
  \includegraphics[width=\linewidth]{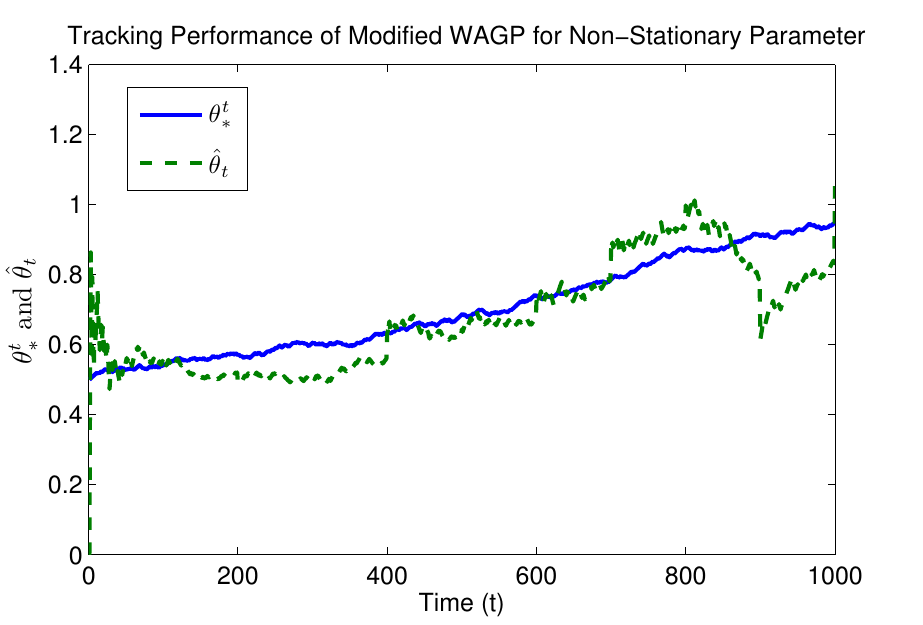}
  \caption{1a : Tracking Performance of Modified WAGP}
  \label{fig:sfig1_nonstationary}
\end{subfigure}%
\begin{subfigure}{.5\textwidth}
  \centering
  \includegraphics[width=\linewidth]{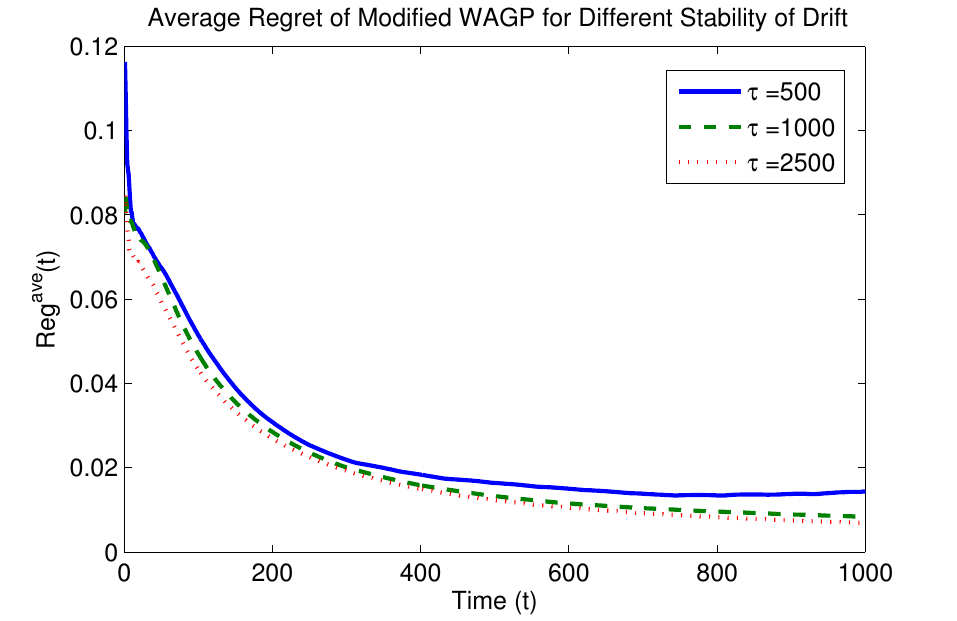}
  \caption{1b : Expected Regret of Modified WAGP}
  \label{fig:sfig2_nonstationary}
\end{subfigure}
\caption{Performance of Modified WAGP for Non-Stationary Global Parameter}
\label{fig:nonstationary}
\end{figure*}

Fig. \ref{fig:nonstationary} illustrates performance of the non-stationary WAGP for the non-stationary dynamic pricing example. We use $\tau = 1000$ to illustrate the tracking performance of the modified WAGP in Fig. \ref{fig:sfig1_nonstationary}. Note that $\tau_h = 100$ for this example. The reward observations used to estimate parameter changes for $t = 200, 300 \ldots, 900$. This results in some jumps in the estimate at these times as seen from Fig. \ref{fig:sfig1_nonstationary}. From this figure it can be seen that  our modified WAGP is able to track the non-stationary global parameter and the slope of the regret is decreasing function of $\tau$ as predicted by Theorem 8. 

\textbf{Experiment 4 (Non-ideal model)}: We show the performance of the WAGP when the revenue of the price $p$ deviates from the expected revenue from the model due to unobserved/unmeasured covariates or unexpected events. Let $R_{p,t} \sim \operatorname{Beta}(1,(1-\tilde{\mu}_p(\theta_{*}))/\tilde{\mu}_p(\theta_{*}))$ where $\tilde{\mu}_p(\theta_{*}) = \mu_p(\theta_{*}) + b_p$ and $b_p \sim \text{Uniform}[-\lambda, \lambda]$ denotes the shift from the model due to some unobserved covariates. Table $4$ shows the regret for different values of $\lambda$ averaged over $100$ different iterations where the model is re-generated in each iteration. As seen from the table, the WAGP outperforms UCB1 and UE algorithms by exploiting (non-ideal) structure in the model.  

\begin{table}[ht]
\label{table:nonideal} 
\centering{}{\fontsize{9}{8}\selectfont %
\begin{tabular}{|l|l|l|l|}
\hline
$\lambda / Algorithm$ & WAGP  & UCB1  & UE     \tabularnewline \hline 
$\lambda = 0.01$  & $1.58$  & $164.85$  & $162.18$   \tabularnewline \hline
$\lambda = 0.05$ & $10.07$ & $169.47$ & $291.40$ \tabularnewline \hline
$\lambda = 0.1$ & $32.68$ & $164.38$ & $326.77$ \tabularnewline \hline
\end{tabular}} 
\protect\caption{Regrets of WAGP, UCB1 and UE on $10000$ samples for different $\lambda$ values}
\end{table} 

\section{Conclusion} \label{sec:conclusion}
In this paper we introduce a new class of MAB problems called Global Bandits (GB). This general class encompasses the previously introduced linearly-parametrized bandits as a special case. We proved that the regret of the GB has three regimes, which we characterized for the regret bound, and showed that the parameter dependent regret is bounded, i.e., it is asymptotically finite. In addition to this, we also proved a parameter-free regret bound, which grows sublinearly over time, where the rate of growth depends on the informativeness of the arms. Future work includes extension of global informativeness to group informativeness, and a foresighted MAB, where the arm selection is based on a foresighted policy that explores the arms according to their level of informativeness rather than the greedy policy.

\section{Appendices}

\subsection{Preliminaries} \label{app:lemmaproofs} 
In all the proofs given below. Let $\boldsymbol{w}(t) := (w_1(t), \ldots, w_K(t))$ be the vector of weights and $\boldsymbol{N}(t) := (N_1(t), \ldots, N_k(t))$ be the vector of counters at time $t$. We have $\boldsymbol{w}(t) =  \boldsymbol{N}(t)/t$. Since $\boldsymbol{N}(t)$ depends on the history, they are both random variables that depend on the sequence of obtained rewards.

\vspace{-0.1in}

\subsection{Proof of Proposition 1} \label{app:prop1proof}  
(i) Let $k$ and $\theta \neq \theta'$ be arbitrary. Then, by Assumption 1, 
$$
|\mu_k(\theta) - \mu_k(\theta')| \geq D_{1,k} |\theta - \theta'|^{\gamma_{1,k}} > 0 
$$
and hence $\mu_k(\theta) \neq \mu_k(\theta')$. 

(ii) Suppose $x = \mu_k(\theta)$ and $x' = \mu_k(\theta')$ for some arbitrary $\theta$ and $\theta'$. Then, by Assumption 1,
$$
|x - x'| \geq D_{1,k} |\mu_k^{-1}(x) - \mu_k^{-1}(x')|^{\gamma_{1,k}}.
$$

\vspace{-0.2in}
\subsection{Preliminary Results}

\begin{lemma} \label{lemma:gap} For the WAGP the following relation between $\hat{\theta}_{t}$
and $\theta_{*}$ holds with probability one: $|\hat{\theta}_{t}-\theta_{*}|\leq\sum_{k=1}^{K}w_{k}(t)\bar{D}_{1}|\hat{X}_{k,t}-\mu_{k}(\theta_{*})|^{\bar{\gamma}_{1}}$.
\end{lemma}

\begin{proof}
Before deriving a bound of gap between the global parameter estimate and true global parameter at time $t$, we let $\tilde{\mu}_k^{-1}(x) = \arg\min_{\theta \in \Theta} |\mu_k(\theta) - x|$. By monotonicity of $\mu_k(\cdot)$ and Proposition 1, we have $|\tilde{\mu}_k^{-1}(x) -\tilde{\mu}_k^{-1}(x')| \leq \bar{D}_1 |x - x'|^{\bar{\gamma}_1}$. Then, 
\begin{eqnarray*}
 |\theta_{*} - \hat{\theta}_t| &=& |\sum_{k=1}^{K} w_k(t)\hat{\theta}_{k,t} -\theta_{*}| = \sum_{k=1}^{K} w_k(t) \left|\theta_{*} - \hat{\theta}_{k,t} \right| \notag \\
 &\leq& \sum_{k=1}^{K} w_k(t)|\tilde{\mu}^{-1}_k(\hat{X}_{k,t}) - \tilde{\mu}^{-1}_k(\tilde{\mu}_k(\theta_{*}))| \notag \\ 
 &\leq& \sum_{k=1}^{K} w_k(t)\bar{D}_1|\hat{X}_{k,t} - \mu_k(\theta_{*})|^{\bar{\gamma}_1}, 
\end{eqnarray*}
where we need to look at following two cases for the first inequality. The first case is $\hat{X}_{k,t} \in {\cal X}_k$ where the statement immediately follows. The second case is $\hat{X}_{k,t} \notin {\cal X}_k$, where the global parameter estimator $\hat{\theta}_{k,t} $ is either $0$ or $1$. 
\end{proof}

\begin{lemma} \label{lemma:onesteploss} The one-step regret of the
WAGP is bounded by $r_{t}(\theta_{*})=\mu^{*}(\theta_{*})-\mu_{I_{t}}(\theta_{*}) \leq 2D_{2}|\theta_{*}-\hat{\theta}_{t-1}|^{\gamma_{2}}$ with probability one, for $t\geq2$. \end{lemma}
\begin{proof}
Note that $I_t \in \argmax_{k \in {\cal K}} \mu_k(\hat{\theta}_{t-1})$. Therefore, we have 
\begin{align}
\mu_{I_t}(\hat{\theta}_{t-1}) - \mu_{k^{*}(\theta_{*})}(\hat{\theta}_{t-1}) \geq 0 \label{eq:add} . 
\end{align}

Since $\mu^{*}(\theta_{*}) = \mu_{k^{*}(\theta_{*})}(\theta_{*})$, we have
\begin{align}
& \mu^{*}(\theta_{*})-\mu_{I_{t}}(\theta_{*})  \notag \\ 
&= \mu_{k^{*}(\theta_{*})}(\theta_{*}) - \mu_{I_{t}}(\theta_{*}) \notag \\ 
& \leq  \mu_{k^{*}(\theta_{*})}(\theta_{*}) - \mu_{I_{t}}(\theta_{*}) + \mu_{I_t}(\hat{\theta}_{t-1}) - \mu_{k^{*}(\theta_{*})}(\hat{\theta}_{t-1}) \notag \\
& = \mu_{k^{*}(\theta_{*})}(\theta_{*})  - \mu_{k^{*}(\theta_{*})}(\hat{\theta}_{t-1}) + \mu_{I_t}(\hat{\theta}_{t-1})- \mu_{I_{t}}(\theta_{*})  \notag \\
& \leq 2D_2|\theta_{*} -\hat{\theta}_{t-1}|^{\gamma_2} , \notag
\end{align}
where the first inequality follows from (\ref{eq:add}) and the second inequality follows from Assumption 1. 
\end{proof}
Let ${\cal G}_{\theta_{*},\hat{\theta}_{t}}(x):=\{|\theta_{*}-\hat{\theta}_{t}|>x\}$
be the event that the distance between the global parameter estimate
and its true value exceeds $x$. Similarly, let ${\cal F}_{\theta_{*},\hat{\theta}_{t}}^{k}(x):=\{|\hat{X}_{k,t}-\mu_{k}(\theta_{*})|>x\}$
be the event that the distance between the sample mean reward estimate
of arm $k$ and the true expected reward of arm $k$ exceeds $x$.
\begin{lemma} \label{lemma:eventbound} For WAGP we have
\begin{align}
{\cal G}_{\theta_{*},\hat{\theta}_{t}}(x) \subseteq \bigcup_{k=1}^{K}{\cal F}_{\theta_{*},\hat{\theta}_{t}}^{k} \left( \left(\frac{x}{\bar{D}_{1} w_k(t) K} \right)^{\frac{1}{\bar{\gamma}_1}} \right)      \notag
\end{align}
with probability one, for $t \geq 2$.
\end{lemma}
\begin{proof}
Observe that
\begin{align}
&\{ |\theta_{*} - \hat{\theta}_t| \leq x \} \notag \\ 
& \supseteq \left\{ \sum_{k=1}^{K} w_k(t) \bar{D}_{1} |\hat{X}_{k,t} - \mu_k(\theta_{*})|^{\bar{\gamma}_{1}} 
\leq x \right\} \notag \\ 
& \supseteq \bigcap_{k=1}^K \left\{ |\hat{X}_{k,t} - \mu_k(\theta_{*}) | \leq \left(\frac{x}{w_k(t) \bar{D}_{1} K}\right)^{1/\bar{\gamma}_{1}}\right\},  \notag
\end{align}
where the first inequality follows from Lemma \ref{lemma:gap}. Then, 
\begin{align}
&\{ |\theta_{*} - \hat{\theta}_t| > x \} \subseteq \notag \\ 
& \;\;\;\;\;\;\;\;\;\;\;\;\; \bigcup_{k=1}^K \left\{ |\hat{X}_{k,t} - \mu_k(\theta_{*}) | > \left(\frac{x}{w_k(t) \bar{D}_{1} K}\right)^{1/\bar{\gamma}_{1}}\right\}. \notag
\end{align}
\end{proof}

\subsection{Proof of Theorem 1} \label{app:theorem2proof} 
 Using Lemma \ref{lemma:gap}, the mean-squared error can be bounded as 
\begin{align}
\notag
& \mathbb{E} \left[ |\theta_{*}-\hat{\theta}_{t}|^{2} \right] \notag \\ 
&\leq \mathbb{E} \left[ \left(\sum_{k=1}^K \bar{D}_{1} w_k(t) |\hat{X}_{k,t} -\mu_k(\theta_{*})|^{\bar{\gamma}_1} \right)^2 \right] \\
& \leq K \bar{D}_{1}^2 \sum_{k=1}^K \mathbb{E} \left[w^2_k(t)|\hat{X}_{k,t}-\mu_k(\theta_{*})|^{2\bar{\gamma}_{1}} \right] , \notag
\end{align}
where the inequality follows from the fact that $\left(\sum_{k=1}^{K} a_k\right)^2 \leq K \sum_{k=1}^K a_k^2$ for any $a_k >0$. Then, 
\begin{align}
\notag
& \mathbb{E} \left[ |\theta_{*}-\hat{\theta}_{t}|^{2} \right] \notag \\ 
& \leq K \bar{D}_{1}^2 \mathbb{E} \left[\sum_{k=1}^{K} w_k^2(t) \mathbb{E} \left[ |\hat{X}_{k,t}-\mu_k(\theta_{*})|^{2\bar{\gamma}_{1}} | \boldsymbol{w}(t) \right] \right] \\ \notag
& \leq K \bar{D}_{1}^2 \mathbb{E} \left[\sum_{k=1}^{K} w_k^2(t) \int_{x=0}^{\infty}\Pr( |\hat{X}_{k,t}-\mu_{k}(\theta_{*})|^{2\bar{\gamma}_{1}} \geq x | \boldsymbol{w}(t)) dx \right] ,  \notag
\end{align}
where the second inequality follows from the fundamental theorem of expectation. Then, we can bound inner expectation as 
\begin{align}
\notag
& \int_{x=0}^{\infty} \Pr( |\hat{X}_{k,t}-\mu_{k}(\theta_{*})|^{2 \bar{\gamma}_{1}} \geq x | \boldsymbol{w}(t)  ) dx \notag \\ 
&\leq \int_{x=0}^{\infty} 2 \exp(- x^{\frac{1}{\bar{\gamma}_{1}}} N_k(t)) \,dx. \notag \\ 
& = 2 \bar{\gamma}_{1} \Gamma(\bar{\gamma}_{1}) N_k(t)^{- \bar{\gamma}_{1}} , \notag
\end{align}
where $\Gamma(\cdot)$ is Gamma function. Then, we have 
\begin{align}
\mathbb{E} [ |\theta_{*}-\hat{\theta}_{t}|^{2}] 
& \leq 2 K \bar{\gamma}_{1} \bar{D}^2_1 \Gamma(\bar{\gamma}_{1}) 
\mathbb{E} \left[\sum_{k=1}^{K}\frac{N_k(t)^{2-\bar{\gamma}_{1}}}{t^2} \right]  \notag \\
& \leq 2 K \bar{\gamma}_{1} \bar{D}^2_1 \Gamma(\bar{\gamma}_{1}) t^{- \bar{\gamma}_{1}} , \notag
\end{align}
where the last inequality follows from the fact that $\mathbb{E} [\sum_{k=1}^{K} N^{2- \bar{\gamma}_{1}}_k(t)/t^2] \leq t^{- \bar{\gamma}_{1}}$ for any $N_k(t)$ since $\sum_{k=1}^{K} N_k(t) =t$ and $\bar{\gamma}_{1} \leq 1$. 
\subsection{Proof of Theorem 2} \label{app:theorem3proof}
By Lemma \ref{lemma:onesteploss} and Jensen's inequality, we have
\begin{align}
\mathbb{E} [r_{t+1}(\theta_{*})] 
\leq 2D_{2} \mathrm{E} \left[ |\theta_{*}-\hat{\theta}_{t}| \right]^{\gamma_{2}} . \label{eq:r_t}
\end{align}
Also by Lemma \ref{lemma:gap} and Jensen's inequality, we have 
\vspace{-0.1in}
\begin{align}
& \mathbb{E} \left[ |\theta_{*}-\hat{\theta}_{t}| \right] \notag \\ 
&  \leq \bar{D}_{1} \mathbb{E} \left[ \sum_{k=1}^{K} w_k(t) \mathbb{E} \left[ |\hat{X}_{k,t} -\mu_k(\theta_{*})|\;| \boldsymbol{w}(t) \right]^{\bar{\gamma}_{1}} \right] , \label{eq:gap}
\end{align}
where $\mathbb{E} [\cdot | \cdot]$ denotes the conditional expectation. Using Hoeffding's inequality, we have for each $k \in {\cal K}$
\begin{align}
&\mathbb{E} \left[|\hat{X}_{k,t}-\mu_{k}(\theta_{*})|\;| \boldsymbol{w}(t) \right] \notag \\ 
 &=\int_{x=0}^{1}\!\text{Pr} \left(|\hat{X}_{k,t}-\mu_{k}(\theta_{*})|>x | \boldsymbol{w}(t) \right) \mathrm{d}x \notag \\
 &\leq \int_{x=0}^{\infty}\!2\exp(-2x^{2}N_{k}(t))\,\mathrm{d}x \leq\sqrt{\frac{\pi}{2N_{k}(t)}} . \label{eq:chernoff1}
\end{align}
Combining (\ref{eq:gap}) and (\ref{eq:chernoff1}), we get
\vspace{-0.1in}
\begin{align}
\mathbb{E} [|\theta_{*} - \hat{\theta}_t|] 
&\leq \bar{D}_{1}(\frac{\pi}{2})^{\frac{\bar{\gamma}_{1}}{2}}\frac{1}{t^{\frac{\bar{\gamma}_{1}}{2}}} 
\mathbb{E} \left[\sum_{k=1}^K {w_k(t)}^{1- \frac{\bar{\gamma}_{1}}{2}} \right]. \label{eq:gap2}
\end{align}
Since $w_k(t) \leq 1$ for all $k \in {\cal K}$, and $\sum_{k=1}^K w_k(t) =1$ for any possible $\boldsymbol{w}(t)$, we have $\mathbb{E}[\sum_{k=1}^{K} w_{k}(t)^{1-\frac{\bar{\gamma}_{1}}{2}}]\leq K^{\frac{\bar{\gamma}_{1}}{2}}$. Then, combining (\ref{eq:r_t}) and (\ref{eq:gap2}), we have
\begin{align}
\mathbb{E} [r_{t+1}(\theta_{*})]\leq  2 \bar{D}_{1}^{\gamma_{2}}D_{2}\frac{\pi}{2}^{\frac{\bar{\gamma}_{1}\gamma_{2}}{2}} K^{\frac{\bar{\gamma}_{1}\gamma_{2}}{2}}\frac{1}{t^{\frac{\bar{\gamma}_{1}\gamma_{2}}{2}}} . \notag
\end{align}
\subsection{Proof of Theorem 3} \label{app:theorem4proof}
This bound is consequence of Theorem 2
and the inequality given in bound where for $\gamma >0$ and $\gamma \neq 1$, $\sum_{t=1}^{T} 1/t^{\gamma} \leq 1 + \frac{(T^{1 - \gamma} - 1)}{1 - \gamma}$, i.e.,
\begin{align}
 \text{Reg}(\theta_{*},T) \leq 2 + \frac{2 \bar{D}_{1}^{\gamma_{2}}D_{2}\frac{\pi}{2}^{\frac{\gamma_{1}\gamma_{2}}{2}}K^{\frac{\bar{\gamma}_{1}\gamma_{2}}{2}}}{1-\frac{\bar{\gamma}_{1}\gamma_{2}}{2}}  T^{1-\frac{\bar{\gamma}_{1}\gamma_{2}}{2}}. \notag
\end{align}

\subsection{Proof of Theorem 4} \label{app:theorem5proof} 
We need to bound the probability of the event that ${I_{t} \not\in \mathcal{K}^{*}(\theta_*)}$. Since at time $t+1$, the arm with the highest $\mu_{k}(\hat{\theta}_{t})$ is selected by the WAGP, $\hat{\theta}_{t}$ should lie in $\Theta\setminus\Theta_{k^{*}(\theta_{*})}$ for a suboptimal arm to be selected. Therefore, we can write,
\begin{align}
& \{{I_{t+1} \not\in \mathcal{K}^{*}(\theta_{*})}\} \notag \\ 
&=\{{\hat{\theta}_{t}\in\Theta\setminus\Theta_{k^{*}(\theta_{*})}}\}\subseteq{{\cal G}_{\theta_{*},\hat{\theta}_{t}}}(\Delta_{*}) . \label{eq:evbound}
\end{align}
By Lemma \ref{lemma:eventbound} and (\ref{eq:evbound}), we have
\begin{align}
 & \Pr(I_{t+1}\not\in \mathcal{K}^{*}(\theta_*)) \notag \\ 
 & \leq\sum_{k=1}^{K} \mathbb{E} \left[ \mathbb{E} \left[ 
 \mathbb{I} \left(  {\cal F}_{\theta_*,\hat{\theta}_{t}}^{k}
 \left( \left( \frac{\Delta_{*}}{w_k(t)\bar{D}_{1} K} \right)^{\frac{1}{\bar{\gamma}_{1}}} \right) 
 \right) | \boldsymbol{N}(t) \right] \right] \notag\\
 & \leq\sum_{k=1}^{K} 2 \mathbb{E} \left[ 
 \exp \left(
 -2 \left( \frac{\Delta_{*}}{w_k(t)\bar{D}_{1} K} \right)^{\frac{2}{\bar{\gamma}_{1}}} w_k(t) t
 \right) 
 \right] \notag\\
 & \leq 2K \exp \left(   
 -2 \left(\frac{\Delta_{*}}{\bar{D}_{1} K }  \right)^{\frac{2}{\bar{\gamma}_{1}}} t 
 \right) ,  \label{eqn:probimportant}
\end{align}
where $\mathbb{I}(\cdot)$ is indicator function which is $1$ if the statement is correct and $0$ otherwise, the first inequality follows from a union bound, the second inequality
is obtained by using the Chernoff-Hoeffding bound, and the last inequality is obtained by
using Lemma \ref{lemma4}. 
We have $\Pr(I_{t+1}\not\in \mathcal{K}^{*}(\theta_{*}))\leq1/t$ for $t>C_{1}(\Delta_{*})$
and $\Pr(I_{t+1} \not\in \mathcal{K}^{*}(\theta_{*}))\leq1/t^{2}$ for $t>C_{2}(\Delta_{*})$.
The bound in the first regime is the result of Theorem \ref{thm:par_indep}.
The bounds in the second and third regimes are obtained by summing the probability given in (\ref{eqn:probimportant})
from $C_{1}(\Delta_{*})$ to $T$ and $C_{2}(\Delta_{*})$ to $T$,
respectively. 

\subsection{Proof of Theorem 5} \label{app:theorem6proof}  
Let $(\Omega,{\cal F},P)$ denote probability space, where $\Omega$ is the sample set and ${\cal F}$ is the $\sigma$-algebra that the probability measure $P$ is defined on. Let $\omega \in \Omega$ denote a sample path. We will prove that there exists event $N\in{\cal F}$ such that $P(N)=0$ and if $\omega \in N^{c}$, then $\lim_{t\rightarrow\infty}I_{t}(\omega) \in \mathcal{K}^{*}(\theta_{*})$.
Define the event ${\cal E}_{t} :=\{I_{t}\neq k^{*}(\theta_{*})\}$.
We show in the proof of Theorem 4 that $\sum_{t=1}^{T}P({\cal E}_{t})<\infty$.
By Borel-Cantelli lemma, we have 
\vspace{-0.1in}
\begin{align}
\Pr({\cal E}_{t}\text{ infinitely often})=\Pr(\limsup_{t\rightarrow\infty}{\cal E}_{t})=0. \notag
\end{align}
Define $N :=\limsup_{t\rightarrow\infty}{\cal E}_{t}$, where $\Pr(N)=0$.
We have, 
\begin{align}
N^{\text{c}}=\liminf_{t\rightarrow\infty}{\cal E}_{t}^{\text{c}}, \notag
\end{align}
where $\Pr(N^{\text{c}})=1-\Pr(N)=1$, which means that $I_{t} \in \mathcal{K}^{*}(\theta_*)$
for all but a finite number of $t$. 
\subsection{Proof of Theorem 6} \label{app:theorem8proof} 
Consider a problem instance with two arms with reward functions $\mu_1(\theta) = \theta^{\gamma}$ and $\mu_2(\theta) = 1 - \theta^{\gamma}$, where $\gamma$ is an odd positive integer and rewards are Bernoulli distributed with $X_{1,t} \sim \text{Ber}(\mu_1(\theta))$ and $X_{2,t} \sim \text{Ber}(\mu_2(\theta))$. Then, optimality regions are $\Theta_1 =[2^{-\frac{1}{\gamma}}, 1]$ and $\Theta_2 = [0, 2^{- \frac{1}{\gamma}}]$. Note that $\gamma_2 =1$ and $\gamma_1 =1/{\gamma}$ for this case. We can show that 
\begin{align}
& |\mu_k(\theta) - \mu_k(\theta')| \leq D_2 | \theta - \theta'| \notag \\ 
& |\mu_k^{-1}(x) - \mu_k^{-1}(x')| \leq \bar{D}_1 |x - x'|^{1/\gamma} \notag 
\end{align}
Let $\theta^{*} = 2^{-\frac{1}{\gamma}}$. Consider following two cases with $\theta_1^{*} = \theta^{*} + \Delta$ and $\theta_2^{*} = \theta^{*} - \Delta$. The optimal arm is $1$ in the first case and $2$ in the second case. In the first case, one step loss due to choosing arm $2$ is lower bounded by 
\begin{align}
& (\theta^{*} +\Delta)^{\gamma} - (1-(\theta^{*}+\Delta)^{\gamma}) \notag \\ 
& = 2 (\theta^{*} +\Delta)^{\gamma}  -1 \notag \\
& = 2 ((\theta^{*})^{\gamma} + {\gamma \choose 1}(\theta^{*})^{\gamma-1} \Delta + {\gamma \choose 2}(\theta^{*})^{\gamma-2} \Delta^2 + \ldots) -1  \notag \\
& \geq 2  {\gamma}2^{\frac{1-\gamma}{\gamma}} \Delta . \notag
\end{align}

Similarly, in the second case, the loss due to choosing arm $1$ is $2  {\gamma} 2^{\frac{1-\gamma}{\gamma}} \Delta + \sum_{i=2}^{\gamma} { \gamma \choose i} (\theta^{*})^{(\gamma-i)} (-\Delta)^i$. Let $A_1(\Delta) = 2  {\gamma} 2^{\frac{1-\gamma}{\gamma}} \Delta + \sum_{i=2}^{\gamma} { \gamma \choose i} (\theta^{*})^{(\gamma-i)} (-\Delta)^i$. 

Define two processes $\nu_1 = \text{Ber}(\mu_1(\theta^{*} + \Delta)) \otimes \text{Ber}(\mu_2(\theta^{*} + \Delta))$ and $\nu_2 = \text{Ber}(\mu_1(\theta^{*} - \Delta)) \otimes \text{Ber}(\mu_2(\theta^{*} - \Delta))$ where $x \otimes y$ denotes the product distribution of $x$ and $y$. Let $\Pr_{\nu}$ denote probability associated with distribution $\nu$. Then, the following holds: 
\begin{align}
&\text{Reg}(\theta^{*}+\Delta, T) + \text{Reg}(\theta_{*} - \Delta, T) \notag \\
& \;\;\;\;\;\; \geq  A_1(\Delta) \sum_{t=1}^{T} \left( \Pr\nolimits_{\nu_1^{\otimes t}}(I_t =2) + \Pr\nolimits_{\nu_2^{\otimes t}}(I_t =1) \right), 
\end{align}
where $\nu^{\otimes t}$ is the $t$ times product distribution of $\nu$. Using well-known lower bounding techniques for the minimax risk of hypothesis testing \cite{tsybakov2009introduction}, we have
\begin{align}
& \text{Reg}(\theta^{*}+\Delta, T) + \text{Reg}(\theta^{*} - \Delta, T) \\
& \;\;\;\;\;\;\;\;\;\; \geq A_1(\Delta) \sum_{t=1}^{T} \exp(- \text{KL}(\nu_1^{\otimes t}, \nu_2^{\otimes t})) , \label{eqn:l4}
\end{align}
where 
\begin{align}
\notag
& \text{KL}(\nu_1^{\otimes t}, \nu_2^{\otimes t}) = t \Big(\text{KL}(\text{Ber}(\mu_1(\theta^{*} + \Delta)), \text{Ber}(\mu_1(\theta^{*} - \Delta)) \notag \\
& \;\;\;\;\;\;\;\;\;\;\;\; +\text{KL}( \text{Ber}(\mu_2(\theta^{*} + \Delta)), \text{Ber}(\mu_2(\theta^{*} - \Delta))\Big) .
\end{align}
Define $A_2 = (1 - \exp(\frac{-4 D_2^2 \Delta^2 T}{(\theta^{*} - \Delta)^{\gamma} (1 - (\theta^{*} - \Delta)^{\gamma} )}))(\theta^{*} - \Delta)^{\gamma} (1 - (\theta^{*} - \Delta)^{\gamma} )$.
By using the fact $\text{KL}(p,q) \leq \frac{(p-q)^2}{q(1-q)}$ \cite{rigollet2010nonparametric}, we can further bound (\ref{eqn:l4}) by
\begin{align}
& \text{Reg}(\theta^{*}+\Delta, T) + \text{Reg}(\theta^{*} - \Delta, T)  \notag \\ 
&\geq A_1(\Delta) \sum_{t=1}^{T}  \exp\left( - \frac{4 D_2 t \Delta^2}{(\theta^{*} - \Delta)^{\gamma}(1 - (\theta^{*} - \Delta)^{\gamma}) } \right)\notag \\
& \geq A_1(\Delta) \frac{ A_2 }{4 D_2 \Delta^2}  \notag 
\end{align}
where $A_2 \in \left(0, 1\right)$ for any $\Delta \in (0, \max(\theta^{*}, 1 -\theta^{*}) )$. Hence, the lower bound for the parameter dependent regret is $\Omega(1)$. In order to show the lower bound for the worst-case regret, observe that
\begin{align}
&\text{Reg}(\theta^{*}+\Delta, T) + \text{Reg}(\theta^{*} - \Delta, T)  \notag \\
&\geq  \frac{2  {\gamma} 2^{\frac{1-\gamma}{\gamma}} A_2}{4 D_2 \Delta}  + \sum_{i=2}^{\gamma} {\gamma \choose i} (-\Delta)^{i-2} (\theta^{*})^{\gamma -i}. \notag
\end{align}
By choosing $\Delta = \frac{1}{\sqrt{T}}$, we can show that for large $T$, $A_2 = 0.25(1 - \exp(-16 D_2^2))$. Hence, worst-case lower bound is $\Omega(\sqrt{T})$.
 
\subsection{Proof of Theorem 7}  \label{app:theorem10proof} 

Without loss of generality, we assume that a unique arm is optimal for $\hat{\theta}_t$ and $\theta_{*}$. First, we show that $ |\hat{\theta}_t - \theta_{*}| = \epsilon$ implies $|\hat{\Delta}_t - \Delta_{*}| \leq \epsilon$.
There are four possible cases for $\hat{\Delta}_t$: 
\begin{itemize}
\item $\theta_{*}$ and $\hat{\theta}_t$ lie in the same optimality interval of the optimal arm, and $\Delta_{*}$ and $\hat{\Delta}_t$ are computed with respect to the same endpoint of that interval. 
\item $\theta_{*}$ and $\hat{\theta}_t$ lie in the same optimality interval and $\Delta_{*}$ and $\hat{\Delta}_t$ are computed with respect to the different endpoints of that interval. 
\item $\theta_{*}$ and $\hat{\theta}_t$ lie in adjacent optimality intervals.
\item $\theta_{*}$ and $\hat{\theta}_t$ lie in non-adjacent optimality intervals.
\end{itemize}

In the first case, $|\hat{\theta}_t - \theta_{*}| = |\hat{\Delta}_t - \Delta_{*}| = \epsilon$. In the second case, $\hat{\Delta}_t$ can not be larger than $\Delta_{*}+ \epsilon$ since in that case $\hat{\theta}_t$ would be computed with respect to the same endpoint of that interval. Similarly, $\hat{\Delta}_t$ can not be smaller than $\Delta_{*}- \epsilon$ since in that case $\theta_{*}$ would be computed with respect to the same endpoint of that interval. In the third and fourth cases, since $|\hat{\theta}_t - \theta_{*}| = \epsilon$, $\hat{\Delta}_t \leq \epsilon - \Delta_{*}$, and hence the difference between $\hat{\Delta}_t$ and $\Delta_{*}$ is smaller than $\epsilon$.

Second, we show that $|\hat{\Delta}_t - \Delta_{*}| < \bar{D}_1 \left(\frac{2 K \log t}{t}\right)^{\frac{\bar{\gamma}_1}{2}}$ holds with high probability. 
\begin{align}
& \Pr\left( |\hat{\Delta}_t - \Delta_{*}| \geq \bar{D}_1 \left(\frac{K \log t}{t}\right)^{\frac{\bar{\gamma}_1}{2}} \right) \notag \\ 
& \leq \Pr\left( |\hat{\theta}_t - \theta_{*}| \geq \bar{D}_1 \left(\frac{K \log t}{t}\right)^{\frac{\bar{\gamma}_1}{2}} \right) \notag \\
& \leq \sum_{k=1}^K 2 \mathbb{E}\left[ \exp\left( -2 \left( \frac{\bar{D}_1 K \left(\frac{ \log t}{t}\right)^{\frac{\bar{\gamma}_1}{2}} }{\bar{D}_1 K w_k(t) }\right)^{\frac{2}{\bar{\gamma}_1}} N_k(t) \right) \Bigg|  N_k(t) \right] \notag \\
&  \leq \sum_{k=1}^K 2 \mathbb{E}\left[ \exp\left( -2 w_k(t)^{1 - \frac{2}{\bar{\gamma}_1}} \log t  \right) \bigg| w_k(t) \right] \notag \\
& \leq 2 K t^{-2}, 
\end{align}
where the second inequality follows from Lemma 3 and Chernoff-Hoeffding inequality and third inequality by Lemma \ref{lemma4}. Then, at time $t$, with probability at least $1 - 2 K t^{-2}$, the following holds: 
\begin{align}
\Delta_{*} - 2 \bar{D}_1 K \left( \frac{\log t}{t} \right)^{\frac{\bar{\gamma}_1}{2}} \leq \tilde{\Delta}_t . \label{eqn:delta_bound}
\end{align}

Also, note that if $t \geq C_2(\Delta_{*}/3)$, then $2 \bar{D}_1 K \left( \frac{ \log t}{t}\right)^{\frac{\bar{\gamma}_1}{2}} \leq \frac{2 \Delta_{*}}{3}$. Thus, for $t \geq C_2(\Delta_{*}/3)$, we have $\Delta_{*}/3 \leq \tilde{\Delta}_t$. 
Note that the BUW follows UCB1 only when $t < C_2(\tilde{\Delta}_t)$. From the above, we know that $C_2(\tilde{\Delta}_t) \leq C_2(\Delta_{*}/3)$ when $t \geq C_2(\Delta_{*}/3)$ with probability at least $1 - 2 K t^{-2}$. This implies that the BUW follows the WAGP with probability at least $1 - 2 K t^{-2}$ when $t \geq C_2(\Delta_{*}/3)$.

We also know from Theorem \ref{thm:par_dep} that the WAGP selects an optimal action with probability at least $1-1/t^2$ when $t > C_2(\Delta_{*})$. Since $C_2(\Delta_{*}/3) > C_2(\Delta_{*})$, when the BUW follows the WAGP, it will select an optimal action with probability at least $1-1/t^2$ when $t > C_2(\Delta_{*}/3)$.

Let $I_t^{g}$ denote the action that selected by algorithm $g \in \{\text{BUW}, \text{WAGP}, \text{UCB1}\}$, $r_t^g(\theta_{*}) = \mu^{*}(\theta_{*}) - \mu_{I_t^{g}}(\theta_{*})$ denote the one-step regret, and $R_{\theta_{*}}^{g}(T_1, T_2)$ denote the cumulative regret incurred by algorithm $g$ from $T_1$ to $T_2$. Then, when $T < C_2(\Delta_{*}/3)$, the regret of the BUW can be written as 
\begin{eqnarray*}
R_{\theta^{*}}^{BUW}(1, T) &\leq& \sum_{t = 1}^T r_t^{UCB1}(\theta_{*}) + 2 K t^{-2} \\
&\leq& R_{\theta^{*}}^{UCB1}(1, T) + \frac{2 K \pi^2}{3}.
\end{eqnarray*}
Moreover, when $T \geq C_2(\Delta_{*}/3)$, we have
\begin{align*}
& R_{\theta^{*}}^{BUW}(C_2(\Delta_{*}/3), T) \\
&\leq \sum_{t = C_2(\Delta_{*}/3)}^{T} r_t^{WAGP}(\theta_{*}) + 2 K t^{-2} \\ 
&\leq R_{\theta^{*}}^{WAGP}(C_2(\Delta_{*}/3), T) + \frac{2 K \pi^2}{3}
\end{align*}
This concludes the parameter-dependent regret bound. 

The worst-case bound can be proven by replacing $\delta_k = \mu^{*} - \mu_k = 1/\sqrt{T K \log T}$ for all $k \not\in \mathcal{K}^{*}(\theta_{*})$ for the regret bound given above.

\subsection{Proof of Theorem 8} \label{app:theorem9proof} 
When the round is clear from the context we use $\hat{\theta}_t$ to represent $\hat{\theta}_{\rho,t}$.
By Lemma \ref{lemma:onesteploss} and Jensen's inequality, we have
\begin{align}
\mathbb{E} \left[ r_{t+1}(\theta_{*}^{t+1}) \right] 
&\leq 2D_{2} \mathbb{E} \left[ |\theta_{*}^{t+1}-\hat{\theta}_{t}| \right]^{\gamma_{2}},  \label{eq:r_t_drift}
\end{align}
where 
$
\hat{\theta}_t = \frac{\sum_{k=1}^{K} N_{k,\rho}(t) \tilde{\mu}_k^{-1}(\hat{X}_{k,\rho,t})}{\tau_{\rho}(t)} , 
$
and $\sum_{k=1}^K N_{k,\rho}(t) = \tau_{\rho}(t)$. Then, by using Lemma \ref{lemma:gap}, we have
\begin{align}
& \mathbb{E} \left[ \left| \hat{\theta}_t - \theta_{*}^{t+1} \right| \right] \notag \\ 
& \leq \frac{\sum_{k=1}^{K} \bar{D}_{1} \mathbb{E} \left[  N_{k,\rho}(t) \mathbb{E} \left[\left|\hat{X}_{k,\rho,t} - \mu_k(\theta_{*}^{t+1})\right| | N_{k,\rho}(t)\right]^{\bar{\gamma}_{1}}  \right]} {\tau_{\rho}(t)}. \notag
\end{align}
Let ${\cal S}_{k,\rho,t}^{\tau_h}$ be the set of times times that arm $k$ is chosen in round $\rho$ by time $t$, i.e., 
\begin{align}
{\cal S}_{k,\rho,t}^{\tau_h} =\{t' \leq t : I_{t'} =k, 2(\rho -1)\tau_h < t' \leq 2 \rho \tau_h \} .      \notag
\end{align}
Clearly, $|{\cal S}_{k,\rho,t}^{\tau_h}| = N_{k,\rho}(t)$. We have, 
\begin{align}
\hat{X}_{k,\rho,t} = \frac{\sum_{t' \in {\cal S}_{k,\rho,t}^{\tau_h}} X_{k,t'}}{N_{k,\rho}(t)} , \notag
\end{align}
where $\mathbb{E}[X_{k,t'}] = \mu_k(\theta_{*}^{t'})$ for all $t' \in {\cal S}_{k,\rho,t}^{\tau_h}$. Define a random variable $\tilde{X}_{k, t'} = X_{k, t'} - \mu_k(\theta_{*}^{t'})$ for all $t' \in {\cal S}_{k,\rho,t}^{\tau_h}$, $k \in {\cal K}$ and $\rho$. Observe that $\{ \tilde{X}_{k, t'}\}_{t' \in {\cal S}_{k,\rho,t}^{\tau_h}}$ is a random sequence with $\mathbb{E}[\tilde{X}_{k, t'}] =0$ and $\tilde{X}_{k, t'} \in [-1,1]$ almost surely for all $k \in {\cal K}$ and $\rho$. Then,
\begin{align}
& \mathbb{E} \left[\left|\hat{X}_{k,\rho, t} - \mu_k(\theta_{*}^{t+1})\right| | N_{k,\rho}(t)\right] \notag \\
&\leq \mathbb{E} \left[ \left| \frac{\sum_{t' \in {\cal S}_{k,\rho,t}^{\tau_h}} (X_{k,t'} - \mu_k(\theta_{*}^{t'}))}{N_{k,\rho}(t)} \right| \right] \notag \\ 
& \;\;\;\;\;\;\;\;\;\;\;\;\; + \frac{\sum_{t' \in {\cal S}_{k,\rho,t}^{\tau_h}} |\mu_k(\theta_{*}^{t'}) - \mu_k(\theta_{*}^{t+1})|}{N_{k,\rho}(t)} \notag \\
& \leq \mathbb{E} \left[ \left| \frac{\sum_{t' \in {\cal S}_{k,\rho,t}^{\tau_h}} \tilde{X}_{k,t'}}{N_{k,\rho}(t)} \right| \right] 
+ \frac{\sum_{t' \in {\cal S}_{k,\rho,t}^{\tau_h}} 2 D_2 |\theta_{*}^{t'} - \theta_{*}^{t+1} |^{\gamma_2}}{N_{k,\rho}(t)} ~, \notag
\end{align}
where for any $t' \in {\cal S}_{k,\rho,t}$, $k \in {\cal K}$ and $\rho$,
\begin{align}
& \mathbb{E} \left[ \left| \frac{\sum_{t' \in {\cal S}_{k,\rho,t}^{\tau_h}} \tilde{X}_{k,t'}}{N_{k,\rho}(t)} \right| \right] \notag \\ 
&= \int_{x=0}^{\infty}\! \Pr\left(\left|\frac{\sum_{t' \in {\cal S}_{k,\rho,t}^{\tau_h}} \tilde{X}_{k,t'}}{N_{k,\rho}(t)}\right| >x\right) \,\mathrm{d}x \notag \\
& \leq \int_{x=0}^{\infty}\! 2 \exp(- x^2 N_{k, \rho}(t)) \,\mathrm{d}x = \sqrt{\frac{\pi}{N_{k,\rho}(t)}} , \label{eqn:shift1}
\end{align}
where the inequality follows from the Chernoff-Hoeffding bound and 
\begin{align}
|\theta_{*}^{t+1} - \theta_{*}^{t'}| \leq ( 2 \tau_h/\tau  ) , \label{eqn:shift2}
\end{align}
since for all $t, t'$ in the same round $|t-t'| \leq 2 \tau_h$. Then, using (\ref{eqn:shift1}) and (\ref{eqn:shift2}), the expected gap between $\theta_{*}^{t+1}$ and $\hat{\theta}_t$ can be bounded as 
\begin{align}
& \mathbb{E} \left[ |\theta_{*}^{t+1} - \hat{\theta}_t | \right] \notag \\ 
& \leq \frac{\sum_{k=1}^{K} \bar{D}_{1} \mathbb{E} \left[ N_{k,\rho}(t)(|\sqrt{\frac{\pi}{N_{k,\rho}(t)}} + 2 D_2 (\frac{2 \tau_h}{\tau})^{\gamma_2} |)^{\bar{\gamma}_{1}} \right] }{\tau_{\rho}(t)} \notag \\ 
& \leq \frac{\sum_{k=1}^{K} \bar{D}_{1} \mathbb{E} \left[ N_{k,\rho}(t)(\frac{\pi}{N_{k,\rho}(t)})^{\frac{\bar{\gamma}_{1}}{2}} \right] }{\tau_{\rho}(t)} \notag \\ 
& \;\;\;\;\;\;\;\;\;\;\;\;\;\;\;\;+ \frac{\sum_{k=1}^{K}  \bar{D}_{1} 2 D_2^{\bar{\gamma}_{1}} ( 2 \tau_h / \tau  )^{ \gamma_2 \bar{\gamma}_{1}} N_{k,\rho}(t)}{\tau_{\rho}(t)} \notag \\ 
& \leq \bar{D}_{1}( (\pi K)^{\frac{\bar{\gamma}_{1}}{2}} \tau_{\rho}(t)^{-\frac{\bar{\gamma}_{1}}{2}} + 2 D_2^{\bar{\gamma}_{1}}
( 2 \tau_h / \tau )^{\bar{\gamma}_{1} \gamma_2} ) \notag \\
& \leq \bar{D}_{1}( (\pi K)^{\frac{\bar{\gamma}_{1}}{2}} \tau_h^{-\frac{\bar{\gamma}_{1}}{2}} + 2 D_2^{\bar{\gamma}_{1}} 
( 2 \tau_h /\tau )^{ \bar{\gamma}_{1} \gamma_2} ) , \notag
\end{align}
where the second inequality follows from the fact that $(a+b)^{\gamma} \leq a^{\gamma} + b^{\gamma}$ for $a,b > 0$ and $0<\gamma \leq 1$, the third inequality is due to the worst case selection process, i.e., $ N_{k,\rho}(t) = \tau_{\rho}(t) / K$ for all $k \in {\cal K}$ where $\tau_{\rho}(t) / K$ is assumed to be integer without loss of generality, and the fourth inequality follows from the fact that $\tau_{\rho}(t) \geq \tau_h$. By choosing $\tau_h = \lceil \tau \rceil^b$, we get the optimal $b = \frac{\gamma_2}{0.5 + \gamma_2}$. Then, cumulative regret at time $T$ can be bounded as

\begin{align}
&\text{Reg}^{\text{ave}}(T)  \notag \\ 
&\leq \tau^{- \frac{ \gamma_2}{0.5 +  \gamma_2}}  + \left( 2 D_2 \bar{D}_{1}^{\gamma_2}[(\pi K)^{\frac{\bar{\gamma}_{1}}{2}} + 2 D_2^{\bar{\gamma}_{1}}] \right)^{\gamma_2} \tau^{- \frac{ \gamma_2^{2} \bar{\gamma}_{1}}{1+ 2  \gamma_2}} , \notag 
\end{align}
which concludes the proof. 
\subsection{Auxiliary Lemma} \label{app:lemma5proof} 
\begin{lemma} \label{lemma4}
For $\gamma <0$, $\delta > 0$, the following bound holds for any $w_k$ with $0 \leq w_k \leq 1$ and $\sum_{k=1}^K w_k =1$: 
$$
\sum_{k=1}^K  \exp(- \delta w_k^{\gamma}) \leq K \exp(- \delta)
$$
\end{lemma}

\begin{proof}
Let $k_{\max} = \argmax_{k} w_k$. Then, 
\begin{eqnarray*}
&&\max_{w_k : \sum_{k=1}^K w_k =1, \; 0 \leq w_k \leq 1} \; \; \sum_{k=1}^K \exp\left( - \delta w_k^{\gamma}\right)  \\ 
& \;& = \max_{w_k : \sum_{k=1}^K w_k =1, \; 0 \leq w_k \leq 1} \; \; \exp\left( \log\left(\sum_{k=1}^K \exp\left( - \delta w_k^{\gamma}\right) \right)\right) \\ 
&\leq&  \max_{w_k : \sum_{k=1}^K w_k =1, \; 0 \leq w_k \leq 1} \;\; \exp \left( \max_{k \in \mathcal{K}} \left( -\delta w_k^{\gamma} \right) + \log K \right) \\ 
&\leq& K \max_{w_k : \sum_{k=1}^K w_k =1, \; 0 \leq w_k \leq 1} \;\; \exp \left( -\delta w_{k_{\max}}^{\gamma}\right) \\
&\leq& K \exp(- \delta).
\end{eqnarray*}
\end{proof}

\bibliographystyle{IEEEtran}
\bibliography{aistats}

\end{document}